%% file: main.tex
\documentclass[runningheads]{llncs}
\usepackage{graphicx}
\usepackage{wrapfig}
\usepackage{booktabs}
\usepackage{multirow}
\usepackage{subcaption}
\usepackage[labelfont=bf,labelsep=period]{caption}
\usepackage{array}
\usepackage{paralist}
\usepackage{comment}
\usepackage{amsmath,amssymb} % define this before the line numbering.
\usepackage{color}
\usepackage{bm}
\usepackage{diagbox}
\usepackage{xspace}
\usepackage{floatrow}
\newfloatcommand{capbtabbox}{table}[][\FBwidth]
\usepackage{float}
\usepackage{algorithm,algpseudocode}
\usepackage{appendix}

\usepackage[width=122mm,left=12mm,paperwidth=146mm,height=193mm,top=12mm,paperheight=217mm]{geometry}
\renewcommand\tabcolsep{3.5pt}

\DeclareMathOperator*{\argmax}{arg\,max}

\newcommand{\sign}{\textup{\textsf{sign}}}

\newcommand{\x}{\bm{x}}
\newcommand{\y}{\bm{y}}

\newcommand{\N}{\mathcal{N}}

\newcommand{\cL}{\mathcal{L}}

\newcommand{\bbB}{\mathbb{B}}
\newcommand{\bbE}{\mathbb{E}}
\newcommand{\bbR}{\mathbb{R}}

\def\etal{\emph{et al}.}

\begin{document}
% \linenumbers
\pagestyle{headings}
\mainmatter
\def\ECCVSubNumber{457}  % Insert your submission number here

\title{Improved Adversarial Training via Learned Optimizer} % Replace with your title

%\begin{comment}
% \titlerunning{ECCV-20 submission ID \ECCVSubNumber} 
% \authorrunning{ECCV-20 submission ID \ECCVSubNumber} 
% \author{Anonymous ECCV submission}
% \institute{Paper ID \ECCVSubNumber}
%\end{comment}
%******************

% \begin{comment}
\titlerunning{}
\author{Yuanhao Xiong\inst{1} \and
Cho-Jui Hsieh\inst{1}}
\authorrunning{Xiong et al.}
\institute{University of California, Los Angeles, CA 90024, USA \\
\email{\{yhxiong,chohsieh\}@cs.ucla.edu}\\
}
% \end{comment}
%******************
\maketitle

\begin{abstract}
Adversarial attack has recently become a tremendous threat to deep learning models. To improve the robustness of machine learning models, adversarial training, formulated as a minimax optimization problem, has been recognized as one of the most effective defense mechanisms. However, the non-convex and non-concave property poses a great challenge to the minimax training. In this paper, we empirically demonstrate that the commonly used PGD attack may not be optimal for inner maximization, and improved inner optimizer can lead to a more robust model. Then we leverage a learning-to-learn~(L2L) framework to train an optimizer with recurrent neural networks, providing update directions and steps adaptively for the inner problem. By co-training optimizer's parameters and model's weights, the proposed framework consistently improves over PGD-based adversarial training and TRADES. 
\end{abstract}

\input{intro.tex}
\input{related.tex}
\input{pre.tex}

\input{l2l.tex}

\input{experiment.tex}

\section{Conclusion}
For defense mechanisms that can be formulated as a minimax optimization problem, 
we propose to replace the inner PGD-based maximizer with a automatically learned RNN maximizer, and show that jointly training the RNN maximizer and classifier can significantly improve the defense performance. Empirical results demonstrate that  the proposed approach can be combined with several minimax defense objectives, including adversarial training and TRADES. 

For future work, it can be a worthwhile direction to address the inadequacy of L2L in dealing with a long-horizon problem. Then we can substitute the learned optimizer for hand-designed algorithms in both inner and outer problems, which enables an entirely automatic process for adversarial training.
\clearpage
\input{ref.tex}
\clearpage
\input{appendix.tex}
\end{document}

%% file: intro.tex
\section{Introduction}
\label{sec:intro}
It has been widely acknowledged that deep neural networks~(DNN) have made tremendous breakthroughs benefiting both academia and industry. Despite being effective, 
many DNN models trained with benign inputs are vulnerable to small and undetectable perturbation added to original data and tend to make wrong predictions under such threats. 
Those perturbed examples, also known as adversarial examples, can be easily constructed by algorithms such as DeepFool~\cite{moosavi2016deepfool}, Fast Gradient Sign Method~(FGSM)~\cite{goodfellow2014explaining},  and Carlini-Wagner~(C\&W) attack~\cite{carlini2017towards}. Moreover, such adversarial attacks can also be conducted in the black-box setting~\cite{brendel2017decision,cheng2018query,cheng2020signopt} and can appear naturally in the physical world~\cite{hendrycks2019natural,kurakin2016adversarial2}. 
This phenomenon can bring about serious consequences in domains such as face recognition and autonomous-driving. Therefore,  how to train a model resistant to adversarial inputs has become an important topic. 

%To fully understand circumstances where DNNs can be fooled easily, researchers have made attempts to gain insights by attacking those models. There is a rapidly growing body of work on manipulating clean examples into adversarial ones. Instances include Fast Gradient Sign Method~(FGSM)~\cite{goodfellow2014explaining}, Projected Gradient Descent~(PGD)~\cite{kurakin2016adversarial}, and Carlini-Wagner~(C\&W) attack~\cite{carlini2017towards} {\color{red}(add deepfool here)}. The fundamental idea of such attack algorithms is to increase the loss value of clean data by using first-order gradient.

A variety of defense methods have been proposed to improve the performance of DNNs against adversarial attacks~\cite{kurakin2016adversarial,samangouei2018defense,wang2019direct,wang2019bilateral,xie2019feature,zhang2019theoretically}. Among them, adversarial training~\cite{kurakin2016adversarial} stands out for its effectiveness. Moreover, \cite{madry2017towards} shows that adversarial training can be formulated as a minimax optimization problem, resembling a game between the attacker and the defender. The formulation is so intuitive that the inner problem aims at generating adversarial examples by maximizing the training loss while the outer one guides the network in the direction that minimizes the loss to resist attacks. 
However, directly obtaining the optimal value of the inner maximization is infeasible, so one has to run an iterative optimization algorithm for a fixed number (often 10) iterations to get an approximate inner maximizer. 
%To solve the inner maximization problem, a commonly used approach is the Projected Gradient (PGD) method.
%{\color{red}(we probably don't want to focus too much on this, since the learned optimizer won't be able to obtain global maxima as well. )}

Existing adversarial training often uses hand-designed general purpose optimizers, such as PGD attack, to (approximately) solve the inner maximization. However, there is an essential property of adversarial training that is rarely explored: 
%Existing approaches usually leverage general optimizers like PGD to solve the inner maximization. Despite being simple, it often converges slowly and may not converge to a good solution. With further in-depth investigation, we observe an important property of this problem that has rarely been explored: 
the maximization problems associated with each sample share very similar structure, and a good inner maximizer for adversarial training only needs to work well for this set of data-dependent problems. To be specific, there are a finite of $n$ maximization problems need to be solved (where $n$ is number of training samples), and those maximization problems share the same objective function along with identical network structure and weights, and the only difference is their input $\x$. Based on this observation, can we have a better optimizer that in particular works well for these very similar and data-dependent problems? 
%As great capability of neural networks has been demonstrated, Learning-to-Learn~(L2L) provides a new direction to find a better solution to the inner problem. L2L is known to be a meta-learning technique that neural networks are utilized for automatic optimization. Is it possible that we learn an optimizer that works well for these similar inner maximization problems?

%{\color{red}Add a paragraph to motivate why we want to use L2L: 1) existing approaches usually use general optimizers to solve the inner maximization. 2) however, they often converge slowly and may not converge to a good maximizer. 3) An important property of this problem that hasn't been explored is that the  maximization problems associated with each sample share very similar structure --- they are using the same loss function with the same network structure and weights, and the only difference is their input $x$. 4) Can we have a better optimizer that works in particular well for these very similar problems?  )}

Motivated by this idea, we propose a learned optimizer for improved adversarial training. Instead of using an existing optimizer with a fixed update rule (such as PGD), we aim at learning the inner maximizer that could be faster and more effective for this particular set of maximization problems.  We have noticed that two works have already put forward algorithms to combine learning to learn with adversarial training~\cite{jiang2018learning,jang2019adversarial}. Both of them adopt a convolutional neural network (CNN) generator to produce malicious perturbations whereas CNN structure might complicate the training process and cannot grasp the essence of the update rule in the long term.  %Promising results have been observed that such learned optimizers beat hand-designed algorithms like SGD in specific tasks~\cite{andrychowicz2016learning,wichrowska2017learned}. 
%{\color{red}Motivated by this idea, we propose XXX. Instead of using a existing optimizer with fixed update rule (such as PGD), we propose to learn the inner maximizer that could be faster and more effective for this particular set of maximization problem. Mention we adopt the L2L framework to model maxmimization as RNN xxx}
 %Given that existing methods to tackle the inner maximization can be regarded as variants of gradient ascent, exploiting L2L to locate a better local maxima for the inner term is prospective. 
%Considering limitations of CNN, 
In contrast, we propose an L2L-based adversarial training method with recurrent neural networks~(RNN). RNN is capable of capturing long-term dependencies and has shown great potentials in predicting update directions and steps adaptively~\cite{lv2017learning}. Thus, following the framework in~\cite{andrychowicz2016learning}, we leverage RNN as the optimizer to generate perturbations in a coordinate-wise manner. Based on the properties of the inner problem, we tailor our RNN optimizer with removed bias and weighted loss for further elaborations to ameliorate issues like short-horizon in L2L~\cite{wu2018understanding}. 

Specifically, our main contributions in this paper are summarized as follows: 
%Compared with previous CNN generators, our method is much easier and more stable to train with less trainable parameters, and can produce stronger attacks iteratively like PGD with its recursive property. 
% \vspace{-2mm}
\begin{itemize}
    \item We first investigate and confirm the improvement in the model robustness from stronger attacks by searching a suitable step size for PGD. 
    \item In replacement of hand-designed algorithms like PGD, an RNN-based optimizer based on the properties of the inner problem is designed to learn a better update rule. In addition to standard adversarial training, the proposed algorithm can also 
    be applied to any other minimax defense objectives such as TRADES~\cite{zhang2019theoretically}.
    %also be applied to the inner maximization in any mini-max adversarial training objective.
    %\item Compared to other learning-to-learn based methods which use the CNN generator, our RNN-based adversarial training sig one is more efficient and stable to train in terms of less trainable parameters and training time.
    \item Comprehensive experimental results show that the proposed method can noticeably improve the robust accuracy of both adversarial training~\cite{madry2017towards} and TRADES~\cite{zhang2019theoretically}. Furthermore, our RNN-based adversarial training significantly outperforms previous CNN-based L2L adversarial training and requires much less number of trainable parameters.
    %    are implemented for general adversarial training as well as TRADES-based adversarial training. Empirical results illustrate the effectiveness of our L2L-based method in both producing stronger attacks and enhancing model robustness.
\end{itemize}

%% file: related.tex
\section{Related Work}
\label{sec:related}

\subsection{Adversarial Attack and Defense}
Model robustness has recently become a great concern for deploying deep learning models in  real-world applications. Goodfellow \etal~\cite{goodfellow2014explaining} succeeded in fooling the model to make wrong predictions by Fast Gradient Sign Method~(FGSM). Subsequently, to produce adversarial examples, IFGSM and Projected Gradient Descent (PGD)~\cite{goodfellow2014explaining,madry2017towards} accumulate attack strength through running FGSM iteratively, and Carlini-Wagner~(C\&W) attack~\cite{carlini2017towards} designs a specific objective function to increase classification errors. Besides these conventional optimization-based methods, there are several algorithms~\cite{reddy2018nag,xiao2018generating} focusing on generating malicious perturbations via neural networks. For instance, Xiao \etal~\cite{xiao2018generating} exploit GAN, which is originally designed for crafting deceptive images, to output corresponding noises added to benign iuput data.
The appearance of various attacks has pushed forward the development of effective defense algorithms to train neural networks that are resistant to adversarial examples. The seminal work of adversarial training has significantly improved adversarial robustness~\cite{madry2017towards}. It has inspired the emergence of various advanced defense algorithms: TRADES~\cite{zhang2019theoretically} is designed to minimize a theoretically-driven upper bound and GAT~\cite{lee2017generative} takes generator-based outputs to train the robust classifier. All these methods can be formulated as a minimax problem~\cite{madry2017towards}, where the defender makes efforts to mitigate negative effects~(outer minimization) brought by adversarial examples from the attacker~(inner maximization). Whereas, performance of such an adversarial game is usually constrained by the quality of solutions to the inner problem~\cite{jang2019adversarial,jiang2018learning}. Intuitively, searching a better maxima for the inner problem can improve the solution of minimax training, leading to improved defensive models.

\subsection{Learning to Learn}
%{\color{red}(Cho: L2L is very broad and includes many subareas. So we need to emphasize that we are only related to the branch that trying to automatically learn an optimizer)}
Recently, learning to learn emerges as a novel technique to efficiently address a variety of problems such as automatic optimization~\cite{andrychowicz2016learning}, few-shot learning~\cite{finn2017model}, and neural architecture search~\cite{elsken2018neural}. In this paper, we emphasize on the subarea of L2L: how to learn an optimizer for better performance. Rather than using human-defined update rules, learning to learn makes use of neural networks for designing optimization algorithms automatically. It is developed originally from \cite{cotter1990fixed} and \cite{younger2001meta}, in which early attempts are made to model adaptive algorithms on simple convex problems. More recently, \cite{andrychowicz2016learning} proposes an LSTM optimizer for some complex optimization problems, such as training a convolutional neural network classifier. Based on this work, elaborations in \cite{lv2017learning} and \cite{wichrowska2017learned} further improve the generalization and scalability for learned optimizers. Moreover, \cite{ruan2019learning} demonstrates that a zeroth order optimizer can also be learned using L2L. 
%first utilizes an L2L framework to generate adversarial examples in black-box attacks defined as zeroth-order optimization, and promising results make it compelling to combine L2L with adversarial training. 
%{\color{red}(which is this line?)}
Potentials of learning-to-learn motivates a line of L2L-based defense which replaces hand-designed methods for solving the inner problem with neural network optimizers. \cite{jiang2018learning} uses a CNN generator mapping clean images into corresponding perturbations. Since it only makes one-step and deterministic attack like FGSM, \cite{jang2019adversarial} modifies the algorithm and produces stronger and more diverse attacks iteratively. Unfortunately, due to the large number of parameters and the lack of ability to capture the long-term dependencies, the CNN generator adds too much difficulty in the optimization, especially for the minimax problem in adversarial training.  Therefore, we adopt an RNN optimizer in our method for a more stable training process as well as a better grasp of the update rule.
% There are many learning-to-learn techniques using in classification tasks. Also some researchers have tried to use L2L in adversarial training with a generator to produce perturbations. ICML will not accept any paper which,
% at the time of submission, is under review for another 

%% file: pre.tex
\section{Preliminaries}
\label{sec:pre}

\subsection{Notations}
We use bold lower-case letters $\x$ and $\y$ to represent clean images and their corresponding labels. An image classification task is considered in this paper with the classifier $f$ parameterized by $\theta$. $\sign(\cdot)$ is an elementwise operation to output the sign of a given input with $\sign(0)=1$. $\bbB(\x, \epsilon)$ denotes the neighborhood of $\x$ as well as the set of admissible perturbed images: $\{\x':\|\x'-\x\|_{\infty}\le\epsilon\}$, where the infinity norm is adopted as the distance metric. We denote by $\Pi$ the projection operator that maps perturbed data to the feasible set. Specifically, $\Pi_{\bbB(\x,\epsilon)}(\x')=\max(\x-\epsilon, \min(\x', \x+\epsilon))$, which is an elementwise operator. $\cL(\cdot,\cdot)$ is a multi-class  loss like cross-entropy.

\subsection{Adversarial Training}
\label{subsec:adv-train}
In this part, we present the formulation of adversarial training, together with some hand-designed optimizers to solve this problem. 
%{\color{red}(I think in this part we are just introducing what is adversarial training. )}
%Our goal is to train a classifier robust to adversarial attacks.
%such as PGD~\cite{madry2017towards} and C\&W~\cite{carlini2017towards}. 
To obtain a robust classifier against adversarial attacks, 
an intuitive idea is to minimize the robust loss, defined as the worst-case loss within a small neighborhood $\bbB(\x,\epsilon)$. 
Adversarial training, which aims to find the weights that minimize the robust loss, can be formulated as a minimax optimization problem in the following way~\cite{madry2017towards}:
\begin{equation}
    \label{eq:adv-train}
\min_{\theta} \bbE_{(\x, \y)\sim D} \left\{\max_{\x'\in\bbB(\x,\epsilon)} \cL(f(\x'),\y)\right\} %{\color{red}\text{(add period after equation if it's the end of sentence. )}}
\end{equation}
% \begin{equation}
%     \label{eq:adv-train}
% \min_{\theta} \bbE \left\{\max_{\x'\in\bbB(\x,\epsilon)} \cL(f(\x'),\y)\right\}
% \end{equation}
where $D$ is the empirical distribution of input data. However,~\eqref{eq:adv-train} only focuses on accuracy over adversarial examples and might cause severe over-fitting issues on the training set. To address this problem, TRADES~\cite{zhang2019theoretically} investigates the trade-off between natural and robust errors and theoretically puts forward a different objective function for adversarial training:
\begin{equation}
\label{eq:trades}
\min_{\theta} \bbE_{(\x, \y)\sim D} \left\{\cL(f(\x),\y)+\hspace{-0.2cm}\max_{\x'\in\bbB(\x,\epsilon)} \cL(f(\x),f(\x'))/\lambda\right\}.
\end{equation}

Note that~\eqref{eq:adv-train} and~\eqref{eq:trades} are both defined as minimax optimization problems, and to solve such saddle point problems, a commonly used approach is to first get an approximate solution $\x'$ of inner maximization based on the current $\theta$, and then use $\x'$ to conduct updates on model weights $\theta$. The adversarial training procedure then iteratively runs this on each batch of samples until convergence.
Clearly, the quality and efficiency of inner maximization is crucial to the performance of adversarial training. The most commonly used inner maximizer is the  projected gradient descent algorithm, which conducts a fixed number of updates: 
%
%Highly nonconcave makes it extremely difficult to directly find the optimal for the inner maximization term, as discussed in \cite{jang2019adversarial,jiang2018learning}. Therefore, researchers turn to approximate the optimal point with hand-designed algorithms. One efficient and effective method to is to make use of the first-order gradient.
\begin{equation}
\label{eq:pgd}
\x_{t+1}' = \Pi_{\bbB(\x,\epsilon)}(\alpha\sign(\nabla_{\x'} \cL(\x_t'))+\x_t').
\end{equation}
Here $\cL(\x_t')$ represents the maximization term in \eqref{eq:adv-train} or \eqref{eq:trades} with abuse of notation. %{\color{red}(We probably haven't defined the projection operator?)}
%After running \eqref{eq:pgd} for a fixed number of iterations, we then use the resulting (approximate) solution to update the classifier. 
%After the ``maxima" is found, we can train the classifier accordingly. 
%Then alternate updates are conducted between adversarial examples and neural network parameters to obtain a robust classifier. 
%{\color{red}(I think we can move 4.1 to 3.3; it's like some preliminary experiments to show it's important to have a better inner maximizer. )}

\subsection{Effects of Adaptive Step Sizes}
\label{sec:bls}
%Currently, many works have proposed to replace hand-designed optimization for inner problem with a learned optimizer, 
We found that the performance of adversarial training crucially depends on the optimization algorithm used for inner maximization, and the current widely used PGD algorithm may not be the optimal choice. 
Here we demonstrate that even a small modification of PGD and without any change to the adversarial training objective can boost the performance of model robustness. 
%
%Given the assumption that adversarial training~(AdvTrain) highly depends on the quality of the local maxima and stronger adversarial examples contribute to model robustness, We conduct a toy experiment for illustration.
We use the CNN structure in \cite{zhang2019theoretically} to train a classifier on MNIST dataset. When 10-step PGD~(denoted by PGD for simplicity) is used for the inner maximization, a constant step size is always adopted, which may not be suitable for the subsequent update. Therefore, we make use of backtracking line search~(BLS) to select a step size adaptively for adversarial training~(AdvTrain as abbreviation). Starting with a maximum candidate step size value $\alpha_0$, we iteratively decrease it by  $\alpha_t=\rho\alpha_{t-1}$ until the following condition is satisfied:
% \begin{equation}
%     \cL(\x'+\alpha_t \bm{p}) \ge \cL(\x')+c\alpha_t\bm{p}^\mathrm{T}\nabla\cL(\x)
% \end{equation}
\begin{equation}
    \cL(\x'+\alpha_t \bm{p}) \ge \cL(\x')+c\alpha_t\bm{p}^\mathrm{T}\bm{p}
\end{equation}
where $\bm{p}=\nabla_{\x'}\cL(\x')$ is a search direction.
%{\color{red}(what is $\x$ in (4)? Isn't that term exactly the same with $\bm{p}$? And what is $c$?)}. 
Based on a selected control parameter $c\in(0,1)$, the condition tests whether the update with step size $\alpha_t$   leads to sufficient increase in the objective function, and it is guaranteed that a sufficiently small $\alpha$ will satisfy the condition so line search will always stop in finite steps. This is standard in gradient ascent (descent) optimization, and see more discussions in \cite{nocedal2006numerical}.
%{\color{red}(cite Stephen wright's numerical optimization book)}. 
%an adequately corresponding increase in the objective function.
%{\color{red}(briefly explain the intuition of this condiction and where does this come from. )}
Following the convention, we set $\rho=0.5$ and $c=10^{-4}$. As shown in Table~\ref{tab:line}, defense with AdvTrain+BLS leads to a more robust model than solving the inner problem only by PGD~($88.71\%$ vs $87.33\%$). At the same time the attacker combined with BLS generates stronger adversarial examples: the robust accuracy of the model trained from vanilla adversarial training drops over $1.2\%$ with PGD+BLS, compared to merely PGD attack. This experiment motivates our efforts to find a better inner maximizer for adversarial training.
%better solutions to the inner problem.
\begin{table}[H]
	\centering
	\caption{Effects of the inner solution quality on robust accuracy (\%)}
	\label{tab:line}
	\centering
	{\footnotesize
	\begin{center}
		\begin{tabular}{c|ccc}
% 			\hline
            \toprule
            \diagbox{Defense}{Attack} & Natural & PGD & PGD+BLS \\
            % \hline
            \midrule
            % Plain &   &  &   \\
            AdvTrain & $96.43$  & $87.33$  & $86.09$   \\
            AdvTrain+BLS & $\mathbf{96.70}$  &  $\mathbf{88.71}$ & $\mathbf{88.00}$ \\
% 			\hline	
            \bottomrule
		\end{tabular}	
	\end{center}}
	%\vspace{-0.2in}
\end{table}

%% file: l2l.tex
\begin{figure}[t]
\centering
\includegraphics[width =\textwidth,height=0.27\textheight,
%trim={0.25in 2.5in 0.4in 2.2in},
clip=false]{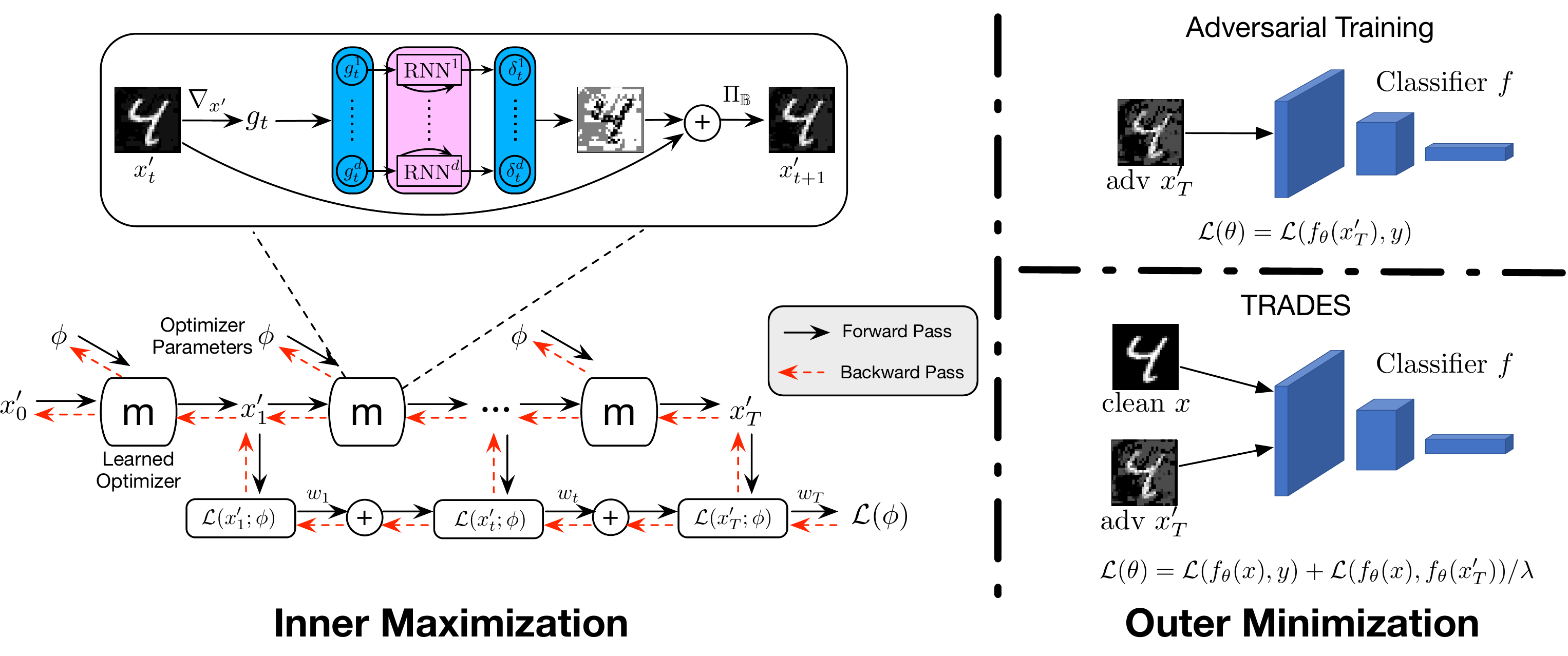}
\caption{Model architecture of our defense method with an RNN optimizer}
\label{fig:arch}
\vspace{-5pt}
\end{figure}
\section{Proposed Algorithm}
\label{sec:l2l}

\subsection{Learning to Learn for Adversarial Training}
%According to the pipeline of adversarial training as well as results of the preliminary experiment in Section~\ref{sec:bls}, 
As mentioned in the previous section, it can be clearly seen that the inner maximizer plays an important role in the performance of adversarial training. 
%the robustness of the classifier. 
However, despite the effectiveness of BLS introduced in Section~\ref{sec:bls}, it is impractical to combine it with adversarial training as multiple line searches %{\color{red}(there is no binary search in line search.)} 
together with loss calculation in this algorithms increase its computational burden significantly. %{\color{red}(briefly mention why there is significant computational overhead.)} 
Then a question arises naturally: is there any automatic way for determining a good step size for inner maximization without too much computation overhead? Moreover, apart from the step size, 
%for much faster convergence and better solutions, 
the question can be extended to whether
such a maximizer can be learned for a particular dataset and model in replacement of a general optimizer like PGD. 
Recently, as a subarea of learning-to-learn, researchers have been investigating whether it is possible to use machine learning, especially neural networks, to learn improved optimizer to replace the hand-designed optimizer~\cite{andrychowicz2016learning,lv2017learning,wichrowska2017learned}. %{\color{red}(cite papers)}. 
%Based on the literature, to learn a general-purpose optimizer is actually nontrivial and unsolved problem. 
However, it is commonly believed that those ML-learned general-purpose optimizers are still not practically useful due to several unsolved issues. For instance, the
exploded gradient~\cite{metz2019understanding} in unrolled optimization impedes generalization of these learned optimizers to longer steps and truncated optimization on the other hand induces short-horizon bias~\cite{wu2018understanding}. %{\color{red}(briefly mention what are those problems)}
%learned optimizers are still 
%Exploded gradient~\cite{metz2019understanding} as well as short-horizon bias~\cite{wu2018understanding} impede development of employing learned optimizers in machine learning tasks like classification.
%Whereas, we show that it is possible to do this for the inner maximization of adversarial training.

In this paper, we show it is possible and practical to learn an optimizer for inner maximization in adversarial training.  
Note that in adversarial training,
the maximization problems share very similar form:  $\max_{\x'\in\bbB(\x,\epsilon)} \cL(f(\x'),\y)$, where they all have  the same loss function $\cL$ and the same network (structure and weights) $f$, 
%
%Note that 
%the maximization problems are quite similar as shown in the equation below:
%\begin{equation}
%    \max_{\x'\in\bbB(\x,\epsilon)} \cL(f(\x'),\y).
%\end{equation}
%We are required to solve this problem with the same loss function $\cL$ and the same network~(structure and weights) $f$, 
and the only difference is their input $\x$ and label $\y$. Furthermore, we only need the maximizer to perform well on a fixed set of $n$ optimization problems for adversarial training. These properties thus enable us to learn a better optimizer that outperforms PGD. 

To allow a learned inner maximizer, we parameterize the learned optimizer by an RNN network. This is following the literature of learning-to-learn~\cite{andrychowicz2016learning}, but we propose several designs as shown below that works better for our inner maximization problem which is a constrained optimization problem instead of a standard unconstrained training task in~\cite{andrychowicz2016learning}. We then jointly optimize the classifier parameters ($\theta$) as well as the parameters of the inner maximizer~($\phi$).  The overall framework can be found in Figure~\ref{fig:arch}. 
Specifically, the inner problem is to maximize vanilla adversarial training loss in~\eqref{eq:adv-train} or TRADES loss in~\eqref{eq:trades}, %denoted by $\cL(\x')$, 
with a constraint that $\x'\in \bbB(\x,\epsilon)$. We expand on adversarial training here and more details about TRADES can be found in Appendix~\ref{appendix:trades}.
%Our goal is to learn an optimizer to obtain a better local maxima for this term. 
%A traditional way to tackle this problem is by Projected Gradient Descent as discussed in Section~\ref{subsec:adv-train}. Now 
With an RNN optimizer $m$ parameterized by $\phi$,
we propose the following parameterized update rule to mimic the PGD update rule in~\eqref{eq:pgd}: 
%the equation in~\eqref{eq:pgd} can be transformed into
% \begin{equation}
% \label{eq:iter}
% \x_{t+1}' = \Pi_{\bbB(\x,\epsilon)}\left(m(\g_t,\h_t)+\x_t'\right)
% \end{equation}
\begin{equation}
 \bm{\delta}_t, \bm{h}_{t+1} =  m_\phi(\bm{g}_t,\bm{h}_t),\quad
 \x_{t+1}' =  \Pi_{\bbB(\x,\epsilon)}\left(\x_t' +\bm{\delta}_t\right). \label{eq:iter}
\end{equation}
% \begin{equation}
%      \x_{t+1}' =  \Pi_{\bbB(\x,\epsilon)}\left(m(\g_t,\h_t)+\x_t'\right) \label{eq:iter}
% \end{equation}
%{\color{red}(you can put (5) and (6) in one line if we run out of space)}
Here, $\bm{g}_t$ is the gradient $\nabla_{\x'}\cL(f(\x'),\y)$ and $\bm{h}_t$ is the hidden state representation. 
%For training stability, the processing function is defined as:
% \begin{equation}
%     \nabla \rightarrow 
%     \begin{cases}
% \left( \frac{\log(\left| \nabla\right|)}{p}, \sign(\nabla) \right)& \text{if} \left| \nabla\right| > e^{-p}\\
% \left(-1, e^p\nabla \right) & \text{otherwise}
% \end{cases}
% \end{equation}
It has to be emphasized that our RNN optimizer generates perturbations coordinate-wisely, in contrast to other L2L based methods which take as input the entire image. This property reduces trainable parameters significantly, making it much easier and faster for training.
% Besides less parameters, coordinate-wisely update enables the optimizer to generalize to adversarial training with different dimensions. For example, the optimizer trained from L\_adv on MNIST can also produce stronger attacks towards classifiers for CIFAR10, and vice versa.
In addition, note that the hidden state of our RNN optimizer plays an important role in the whole optimization. A separate hidden state for each coordinate guarantees the different update behavior. And it contains richer information like the trajectory of loss gradients mentioned in~\cite{jang2019adversarial} but can produce a recursive update with a simpler structure.

For the RNN design, we mainly follow the structure in~\cite{andrychowicz2016learning} but with some modifications to make it more suitable to adversarial training. We can expand the computation of perturbation %{\color{red}(what noise? Should be the update rule of $m$?)} 
for each step as:
\begin{align}
 \bm{\delta}_t &=  \text{tanh}(\bm{Vh}_t + \bm{b}_1), \label{eq:delta} \\
 \bm{h}_{t+1} &= \text{tanh}(\bm{Ug}_t + \bm{Wh}_t + \bm{b}_2) \label{eq:h}
\end{align}
where $\bm{h}_t\in \bbR^d$, $\bm{V}\in\bbR^{1\times d}$, $\bm{U}\in \bbR^{d\times 1}$, $\bm{W}\in \bbR^{d\times d}$, $\bm{b}_1 \in \bbR$ and $\bm{b}_2 \in \bbR^d$ in the coordinate-wise update manner. As the optimization proceeds, the gradient will become much smaller when approaching the local maxima. At that time, a stable value of the perturbation is expected without much change between two consecutive iterations. However, from \eqref{eq:delta} and \eqref{eq:h}, we can clearly see that despite small $\bm{g}_t$, the update rule will still produce an update with magnitude proportional to $\text{tanh}(\bm{b}_1)$. Imagine the case where the exact optimal value is found with an all-zero hidden state~($\bm{b}_2$ needs to be zero as well), $\bm{\delta}_t=\text{tanh}(\bm{b}_1)$ with a non-zero bias will push the adversarial example away from the optimal one. 
Thus, two bias terms $\bm{b}_1$ and $\bm{b}_2$ are problematic for optimization close to the optimal solution. Due to the short horizon of the inner maximization in adversarial training, it is unlikely for the network to learn zero bias terms. Therefore, to ensure stable training, we remove the bias terms in the vanilla RNN in all implementations. 

% For the structure of RNN, we do not use the LSTM. Instead, we design it for the .xxx
With an L2L framework, we simultaneously train the RNN optimizer parameters $\phi$ and the classifier weights $\theta$ together. The joint optimization problem can be formulated as follows:
    % \begin{equation}
    %     \min_w  
    %     \max_\phi\bbE_{(\x, \y)\sim D}\left\{\cL(v_{m_{\phi}}(\x), \y)\right\}
    % \end{equation}
\begin{align}
\min_\theta \ & \  \bbE_{(\x, \y)\sim D}\left\{\cL(f_\theta(\x'_T(\phi^*)), \y)\right\}
 \\
\text{s.t. }  & \ \phi^* = \argmax\cL(\phi) 
%= \sum_{t=1}^{T}w_t\cL(\x'_t;\phi)
\end{align}
    %{\color{red}(Maybe rewrite all the loss as $\bbE_{(\x, \y)\sim D}$ where $D$ is the empirical distribution)}
    %{\color{red}(I think $\max_\phi$ is outside expectation since we use the same RNN for all samples.)}
where $\x'_T(\phi^*)$ is computed by running Eq.\eqref{eq:iter} $T$ times iteratively. Since the learned optimizer aims at finding a better solution to the inner maximization term, the objective function for training it in the horizon $T$ is defined as:
\begin{equation}
\label{eq:loss}
    \cL(\phi) = \sum_{t=1}^{T}w_t\cL(f_\theta(\x'_t(\phi)),\y).
\end{equation}
%{\color{red}(not clear, $\cL(\phi)$ doesn't appear in eq(9).)}
Note that if we set $w_t=0$ for all $t<T$ and $w_T=1$, then \eqref{eq:loss} implies that our learned maximizer $m_\phi$ will maximize the loss after $T$ iterations. However, in practice we found that considering intermediate iterations can further improve the performance since it will make the maximizer converges faster even after conducting one or few iterations.  Therefore in the experiments we set an increased weights $w_t=t$ for $t=1, \dots, T$. Note that \cite{metz2019understanding} showed that this kind of unrolled optimization may lead to some issues such as exploded gradients which is still an unsolved problem in L2L. However, in adversarial training we only need to set a relative small $T$ (e.g., $T=10$) so we do not encounter that issue. 
%Given the objective in \eqref{eq:loss}, updating parameters of the RNN optimizer $\phi$ is usually achieved by unrolled optimization combined with truncated Backpropagation Through Time~(BPTT)~\cite{andrychowicz2016learning}.
%
%But \cite{metz2019understanding} have pointed out that the truncation may induce some issues like increasing bias with truncated gradients. Again the property of short horizon~(10 steps in our paper) for maximization in adversarial training makes it unnecessary to use truncated BPTT. Then $\phi$ can be trained by unrolling the whole optimization steps without too gradient explosion. Furthermore, to make BPTT more efficient, we exploit the weighted sum of the loss value for each time step. Linearly increasing weight (e.g.,$w_t=t$) is preferred to guide the training in the direction where more attention is paid to the final loss instead of the initial stage.

% espite the fact that learning-to-learn has proved successful in various tasks like classification, there are still some problems influencing applications of this technique. A fundamental one lies in the unrolled optimization of learning to learn. Unrolled optimization. Despite the fact that learning-to-learn has proved successful in various tasks like classification, there are still some problems influencing applications of this technique. A fundamental one lies in the unrolled optimization of learning to learn.

While updating the learned optimizer, corresponding adversarial examples are produced together. We can then train the classifier by minimizing the loss accordingly. The whole algorithm is presented in Algorithm~\ref{alg:rnn-adv}.

% \begin{algorithm}[!htb]
%   \caption{RNN-based adversarial training}
%   \label{alg:l2l-adv}
% \begin{algorithmic}
%   \STATE {\bfseries Input:} clean data $\{(\x_i,y_i)\}$, clean data $\{(\x_i,y_i)\}$, clean data $\{(\x_i,y_i)\}$, clean data $\{(\x_i,y_i)\}$, 
%   \REPEAT
%   \STATE Sample mini-batch $B$ from clean data.
%   \FOR{$(\x_i, y_i)$ {\bfseries in} $B$}
%   \FOR{$t=1,\dots, T$}
% %   \IF{$x_i > x_{i+1}$}
% %   \STATE $g_t=\nabla_{\mathbf{x^'}_i}$
%   \STATE $\g_t \leftarrow \text{Preprocess}(\nabla_{\x_i'}\cL(\x_{i,t}'))$
%   \STATE ${\delta}_t,\h_{t+1} \leftarrow m_{\phi}(\g_t, \h_t)$
% %   \STATE $\x_i'=\Pi_{X}(\mathbf{x}_i'+\delta_t)$
%     \STATE Coordinate-wisely update
%     \STATE $\x_i'\leftarrow \Pi_{X}(\mathbf{x}_i'+\delta_t)$
%     \STATE $\cL_{\phi} \leftarrow \cL_{\phi} + w_t\cL(\x') $
% %   \ENDIF
%   \ENDFOR
%   \STATE $\cL_{\theta} \leftarrow \cL+\cL(\x')$ or $\cL_{\theta} \leftarrow \cL(f(\x),y)$ 
%   \ENDFOR
%   \STATE Update $\phi$ over $L_\phi=$
%   \STATE Updata $\theta$ over $L_\theta=$
%   \UNTIL{$noChange$ is $true$}
% \end{algorithmic}
% \end{algorithm}

% \subsection{Unrolled Optimization of Attack}
% Despite the fact that learning-to-learn has proved successful in various tasks like classification, there are still some problems influencing applications of this technique. A fundamental one lies in the unrolled optimization of learning to learn. 

% \subsection{Coordinatewisely Update}
% Previous methods have proposed to use a CNN generator to produce perturbations in adversarial training. However, 
\begin{algorithm}[th]
 \caption{RNN-based adversarial training}\label{alg:rnn-adv}
 \begin{algorithmic}[1]
 \State \textbf{Input}: clean data $\{(\x,\y)\}$, batch size $B$, step sizes $\alpha_1$ and $\alpha_2$, number of inner iterations $T$, classifier parameterized by $\theta$, RNN optimizer parameterized by $\phi$
 \State \textbf{Output}: Robust classifer $f_\theta$, learned optimizer $m_\phi$
 \State Randomly initialize $f_\theta$ and $m_\phi$, or initialize them with pre-trained configurations
 \Repeat
   \State Sample a mini-batch $M$ from clean data.
   \For{$(\x, \y)$ {\bfseries in} $B$}
   \State Initialization: $\bm{h}_0\leftarrow0$, $\cL_{\theta}\leftarrow 0$, $\cL_{\phi}\leftarrow 0$
   \State Gaussian augmentation: $\x_0' \leftarrow \x + 0.001\cdot\N(\bm{0},\bm{I})$
   \For{$t=0,\dots, T-1$} 
  \State $\bm{g}_t \leftarrow \nabla_{\x'}\cL(f_\theta(\x_{t}'),\y)$
  \State ${\bm{\delta}}_t,\bm{h}_{t+1} \leftarrow m_{\phi}(\bm{g}_t, \bm{h}_{t})$, where coordinate-wise update is applied
    \State $\x_{t+1}'\leftarrow \Pi_{\bbB(\x,\epsilon)}(\x_{t}'+\bm{\delta}_t)$
    \State $\cL_{\phi} \leftarrow \cL_{\phi} + w_{t+1}\cL(f_\theta(\x'_{t+1}),\y) $, where $w_{t+1}=t+1$ 
%   \ENDIF
   \EndFor
   \State $\cL_{\theta} \leftarrow \cL_{\theta}+\cL(f_\theta(\x_T'),\y)$ 
   \EndFor
   \State Update $\phi$ by $\phi \leftarrow \phi+\alpha_{1}\nabla_\phi\cL_{\phi}/B$
   \State Update $\theta$ by $\theta \leftarrow \theta - \alpha_{2}\nabla_\theta\cL_{\theta}/B$
 \Until{training converged}
 \end{algorithmic}

\end{algorithm}
\subsection{Advantages over Other L2L-based Methods}

Previous methods have proposed to use a CNN generator~\cite{jang2019adversarial,jiang2018learning} to produce perturbations in adversarial training. %comparing it with generative adversarial training (GAN) {\color{red}(you mean they compare with GAN in their paper? Why do we need to mention this? }. 
%{\color{red}(mention what's the difference between [12,13] and ours.)} 
However, CNN-based generator has a larger number of trainable parameters, which makes it hard to train. In Table~\ref{tab:time}, the detailed properties including the number of parameter and training time per epoch are provided for different learning-to-learn based methods. We can observe that our proposed RNN approach stands out with the smallest parameters as well as efficiency in training. Specifically, our RNN optimizer only has $120$ parameters, almost 5000 times fewer than L2LDA while the training time per epoch is 268.50s~(RNN-TRADES %{\color{red}(RNN-TRADES?)} 
only consumes 443.52s per training epoch) v.s. 1972.41s. 
Furthermore, our method also leads to better empirical performance, as shown in our main comparison in Table~\ref{tab:mnist},~\ref{tab:vgg} and~\ref{tab:wide}. Comparison of our variants and original adversarial training methods can be found in Appendix \ref{appendix:analysis}.
%Table~, the detailed properties including the number of parameter and training time per epoch are provided for different learning-to-learn based methods. We can observe that our proposed one stands out with smallest parameters as well as efficiency in training.

% Previous methods have proposed to use a CNN generator to produce perturbations in adversarial training, comparing it with generative adversarial training (GAN). However, CNN-based generator has a larger number of parameters and is hard to train. In Table~, the detailed properties including the number of parameter and training time per epoch are provided for different learning-to-learn based methods. We can observe that our proposed one stands out with smallest parameters as well as efficiency in training.
% \begin{table}[htb!]
% 	\centering
% 	\caption{Comparion among different L2L-based methods}
% 	\centering
% 	{\footnotesize
% 	\begin{center}
% 		\begin{tabular}{ccc}
% 			\toprule
%              & RNN &  L2LDA \\
%             \midrule
%             % Plain &   &  &   \\
%             number of parameters   & 120  & 500944     \\
%             % \hline
%             \midrule
%             Traning time per epoch   &  & \\
% % 			\hline	
% \bottomrule
% 		\end{tabular}	
% 	\end{center}}
% 	%\vspace{-0.2in}
% \end{table}

\begin{table}[htb!]
	\centering
	\caption{Comparion among different L2L-based methods}
	 \label{tab:time}
	\centering
	{\footnotesize
	\begin{center}
		\begin{tabular}{c|c|c}
			\toprule
             & Number of parameters &  Training time per epoch (s)\\
            \midrule
            % Plain &   &  &   \\
             RNN-Adv  & \textbf{120}  & \textbf{268.50}      \\
             \midrule
             RNN-TRADES & \textbf{120} & 443.52 \\
            % \hline
            \midrule
             L2LDA  & 500944  & 1972.41 \\
% 			\hline	
\bottomrule
		\end{tabular}	
	\end{center}}
	%\vspace{-0.2in}
\end{table}

%% file: experiment.tex
\section{Experimental Results}
\label{sec:exp}
In this section, we present experimental results of our proposed RNN-based adversarial training. 
We compare our method with various baselines against  both white-box and black-box attack. 
In addition, different datasets and network architectures are also evaluated.
%{\color{red}(We want to send the message that the proposed method is a better optimization framework for solving various existing adversarial training formulations, and can be combined with any min-max objective. )}

\subsection{Experimental Settings}
\begin{itemize}
\item 
\textbf{Datasets and classifier networks.} We mainly use MNIST
~\cite{lecun1998gradient} and CIFAR-10~\cite{krizhevsky2010cifar} datasets for performance evaluation in our experiments. For MNIST, the CNN architecture with four convolutional layers in \cite{carlini2017towards} is adopted as the classifier. For CIFAR-10, we use both the standard VGG-16~\cite{simonyan2014very} and Wide ResNet~\cite{zagoruyko2016wide},
which has been used in most of the previous defense papers including adversarial training~\cite{madry2017towards} and TRADES~\cite{zhang2019theoretically}. We also conduct an additional experiment on Restricted ImageNet~\cite{tsipras2018robustness} with ResNet-18 and results are presented in Appendix~\ref{appendix:imagenet}.
%{\color{red} (also talk about Restrcted ImageNet and say the results are in the appendix?)}

\item
\textbf{Baselines for Comparison.}
%{\color{red}(here we can say our method is an optimization framework that is irrelevant to what min-max objective function is used. So we choose two most popular min-max formulation (adv-train and TRADES) and replace their original PGD-based optimization by the proposed L2L-based optimization. )}
Note that our method is an optimization framework which is irrelevant to what minimax objective function is used. Therefore we choose two most popular minimax formulations, AdvTrain\footnote{https://github.com/xuanqing94/BayesianDefense}~\cite{madry2017towards} and TRADES\footnote{https://github.com/yaodongyu/TRADES}~\cite{zhang2019theoretically}, and substitute the proposed L2L-based optimization for their original PGD-based algorithm.
%{\color{red} (maybe don't mention this since people will ask us to compare. In fact we should probably try to apply our method to the SOTA formulation (e.g., bilateral or minhao's algorithm) if we have time)}. 
Moreover, we also compare with a previous L2L defense mechanism L2LDA\footnote{https://github.com/YunseokJANG/l2l-da}~\cite{jang2019adversarial} which outperforms other L2L-based methods for thorough comparison. 
%{\color{red}(maybe mention that L2LDA has shown outperforming the other L2L defense paper?)} 
We use the source code provided by the authors on github with their recommended hyper-parameters for all these baseline methods. 
%All these methods are implemented from open-source repositories with their recommended hyper-parameters. 

\item
\textbf{Evaluation and implementation details.} Defense algorithms are usually evaluated by classification accuracy under different attacks.  Effective attack algorithms including PGD, C\&W and the attacker of L2LDA are used for evaluating the model robustness, with the maximum $\ell_\infty$ perturbation strength $\epsilon=0.3$ for MNIST and $\epsilon=8/255$ for CIFAR-10. For PGD, we run 10 and 100 iterations~(PGD-10 and -100) with the step size $\eta=\epsilon/4$, as suggested in~\cite{jang2019adversarial}. C\&W is implemented with 100 iterations in the infinity norm. For L2LDA attacker, it is learned from L2LDA~\cite{jang2019adversarial} under different settings with 10 attack
steps. In addition, we also uses the learned optimizer of RNN-Adv to conduct 10-step attacks.
%{\color{red}mention RLL-Adv attacker and say it's also 10 steps(?)}

\end{itemize}
%{\color{red}(did we mention that all the adversarial training methods are using 10 inner iterations?)}
For our proposed RNN-based defense, we use a one-layer vanilla RNN with the hidden size of 10 as the optimizer for the inner maximization. Since we test our method under two different minimax losses, we name them as RNN-Adv and RNN-TRADES respectively. The classifier and the optimizer are updated alternately according to the Algorithm~\ref{alg:rnn-adv}.  All algorithms are implemented in PyTorch-1.1.0 with four NVIDIA 1080Ti GPUs. Note that all adversarial training methods adopt 10-step inner optimization for fair comparison. We run each defense method five times with different random seeds and report the lowest classification accuracy. 
%We use SGD with learning rate xx and Adam with learning rate.

\subsection{Performance on White-box Attacks}
\label{subsec:white}
We demonstrate the robustness of models trained from different defense methods under the white-box setting in this part. Experimental results are shown in Table~\ref{tab:mnist},~\ref{tab:vgg} and \ref{tab:wide}. From these three tables, we can observe that our proposed L2L-based adversarial training with RNN always outperforms its counterparts.
\begin{table}[!htb]
	\centering
	\caption{Robust accuracy under white-box attacks (MNIST, 4-layer CNN)}
    \label{tab:mnist}
	\centering
	{\footnotesize
	\begin{center}
		\begin{tabular}{c|cccccc||c}
% 			\hline
\toprule
            \diagbox{Defense}{Attack} & Natural &  PGD-10 & PGD-100 & CW100 & L2LDA & RNN-Adv & Min \\
            % \hline
            \midrule
            Plain & 99.46 &   1.04 & 0.42  & 83.63 & 5.94 & 0.79 & 0.42  \\
            \midrule
            AdvTrain & 99.17 &  94.89 & 94.28 & 98.38 &95.83 & 94.39 & 94.28  \\
            % CW100 & & & & & &  \\
            %TRADES & 99.52  & & 95.77 & & & 95.64 & \\
            TRADES & \textbf{99.52}   & 95.77 & 95.50 &98.72 & 96.03 & 95.50 & 95.50 \\
            \midrule
            L2LDA & 98.76  &94.73 &93.22 & 97.69 & 95.28 & 93.16 & 93.16 \\
            \midrule
            RNN-Adv & 99.20  & 95.80 & 95.62& 98.75 & 96.05 & 95.51 & 95.51  \\
            RNN-TRADES &99.46  & \textbf{96.09} & \textbf{95.83} & \textbf{98.85} & \textbf{96.56} & \textbf{95.80} & \textbf{95.80} \\
           % LSTM-Trades & & & & & & & \\
% 			\hline
\bottomrule
		\end{tabular}	
	\end{center}}
	%\vspace{-0.2in}
\end{table}

\begin{table}[htb!]
	\centering
	\caption{Robust accuracy under white-box attacks (CIFAR-10, VGG-16)}
	\label{tab:vgg}
	\centering
	{\footnotesize
	\begin{center}
		\begin{tabular}{c|cccccc||c}
% 			\hline
\toprule
            \diagbox{Defense}{Attack} & Natural &  PGD-10 & PGD-100 & CW100 & L2LDA & RNN-Adv & Min \\
            % \hline
            \midrule
            Plain & \textbf{93.66}  & 0.74 & 0.09 & 0.08 & 0.89 & 0.43 & 0.08 \\
            \midrule
            AdvTrain & 81.11  & 42.32 & 40.75  &42.26  &43.55 &41.07 & 40.75  \\
            % CW100 & & & & & &  \\
            %TRADES & 99.52  & & 95.77 & & & 95.64 & \\
            TRADES & 78.08   & 48.83 & 48.30  &45.94 & 49.94 & 48.38 & 45.95 \\
            \midrule
            L2LDA & 77.47  & 35.49 & 34.27 &35.31 & 36.27 &  34.54 & 34.27\\
            \midrule
            RNN-Adv & 81.22  & 44.98 & 42.89 & 43.67 & 46.20 & 43.21 & 42.89  \\
            RNN-TRADES &80.76  & \textbf{50.23} & \textbf{49.42} & \textbf{47.23} & \textbf{51.29} & \textbf{49.49} & \textbf{47.23} \\

           % LSTM-Trades & & & & & & & \\
% 			\hline	
\bottomrule
		\end{tabular}	
	\end{center}}
	%\vspace{-0.2in}
\end{table}

\begin{table}[htb!]
	\centering
	\caption{Robust accuracy under white-box attakcs (CIFAR-10, WideResNet)}
	\label{tab:wide}
	\centering
	{\footnotesize
	\begin{center}
		\begin{tabular}{c|cccccc||c}
		
% 			\hline
\toprule
            \diagbox{Defense}{Attack} & Natural & PGD-10 & PGD-100 & CW100 & L2LDA & RNN-Adv & Min\\
            % \hline
            \midrule
            Plain & \textbf{95.14}  & 0.01 & 0.00   & 0.00 &0.02 &0.00 & 0.00 \\
            \midrule
            AdvTrain & 86.28 & 46.64 & 45.13   &46.64 &48.46  &45.41 & 45.13 \\
            %TRADES & 99.52  & & 95.77 & & & 95.64 & \\
            TRADES & 85.89 & 54.28 &52.68 &53.68 &56.49 & 53.00 & 52.68  \\
            \midrule
            L2LDA & 85.30 &45.47 & 44.35  & 44.19& 47.16& 44.54 & 44.19  \\
            \midrule
            RNN-Adv & 85.92  &47.62 & 45.98   &47.26 &49.40 & 46.23 & 45.98  \\
            RNN-TRADES  & 84.21 & \textbf{56.35} & \textbf{55.68} & \textbf{54.11} & \textbf{58.86} & \textbf{55.80} & \textbf{54.11} \\
% 			\hline
\bottomrule
		\end{tabular}	
	\end{center}}
	%\vspace{-0.2in}
\end{table}
To be specific, our method achieves $95.80\%$ robust accuracy among various attacks on MNIST dataset. On CIFAR-10, RNN-TRADES reaches $47.23\%$ and $54.11$ for VGG-16 and Wide ResNet with $1.28\%$ and $1.43\%$ gain over other baselines. It should be stressed that our method surpasses L2LDA (the previous CNN-based L2L method) noticeably. For conventional defense algorithms, our L2L-based variant improves the original method by $1\%-2\%$ percents under different attacks from comparison of robust accuracy in AdvTrain and RNN-Adv. A similar phenomenon can also be observed in TRADES and RNN-TRADES. Since previous works of L2L-based defense only concentrate on PGD-based adversarial training, the substantial performance gain indicates that the learned optimizer can contribute to the minimax problem in TRADES as well. Furthermore, apart from traditional attack algorithms, we leverage our RNN optimizer learned from adversarial training as the attacker~(the column RNN-Adv). Results in three experiments show that compared with other general attackers when conducting 10 iterations such as PGD-10 and L2LDA, ours is capable of producing much stronger perturbations which lead to low
robust accuracy.
%{\color{red}(but ours is not as good compared with PGD100 and CW100, maybe say our attacker is better than other methods when conducting 10 iterations. )} {\color{blue}(is it necessary to say that? We do not conduct 10-step CW attack at all.)}

\subsection{Analysis}

\begin{figure}[t] 
    \centering
    \begin{subfigure}[ht]{0.45\textwidth}
        \centering
        \includegraphics[width=\textwidth,
        trim={0.4in 0in 0in 0in},
        clip=false]{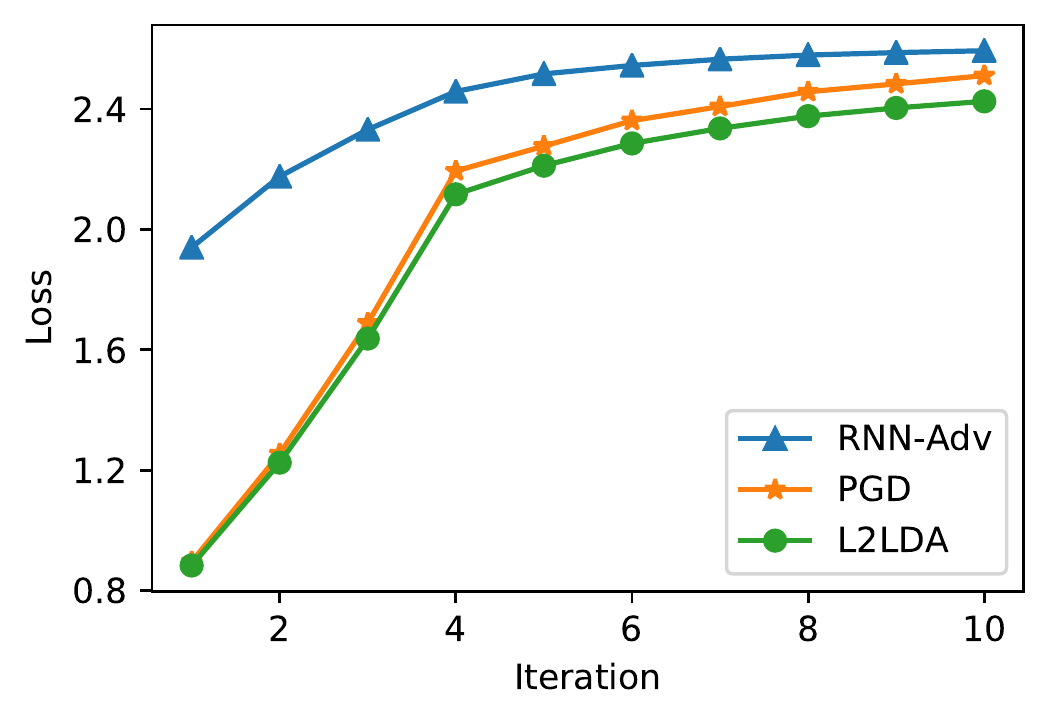}
        \caption{AdvTrain} 
        \label{fig:bianryloss}
    \end{subfigure}
    \begin{subfigure}[ht]{0.45\textwidth}
        \centering
        \includegraphics[width=\textwidth, 
        trim={0in 0in 0.4in 0in},
        clip=false]{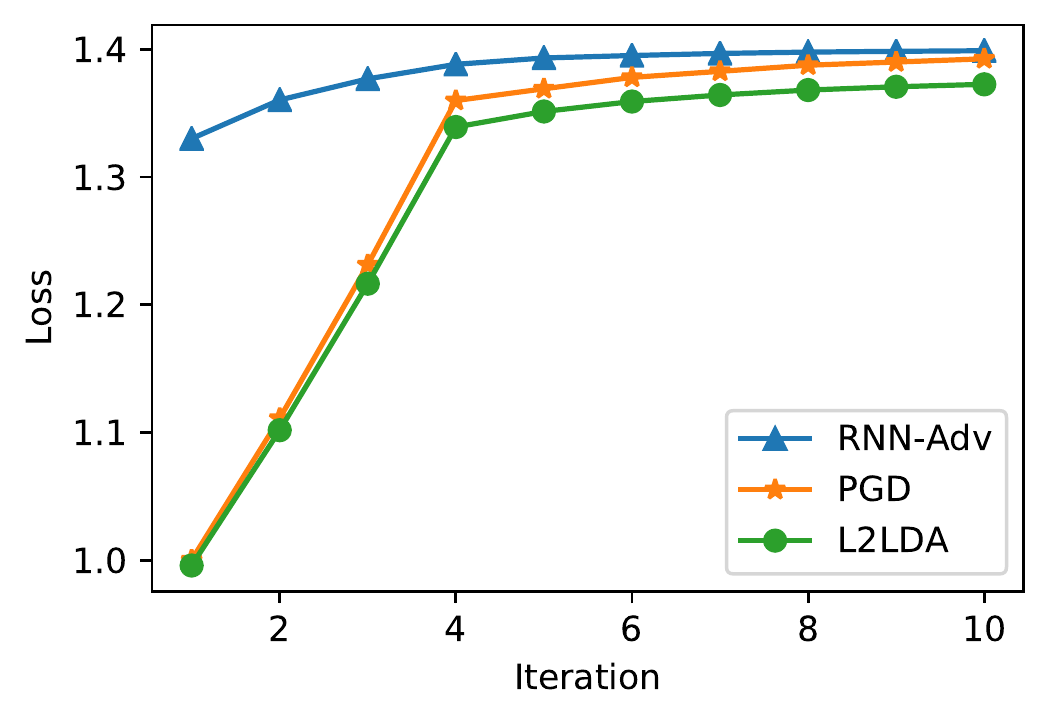}
        \caption{TRADES} 
        \label{fig:bianrycontrolq}
    \end{subfigure}
    \begin{subfigure}[ht]{0.45\textwidth}
        \centering
        \includegraphics[width=\textwidth, 
        trim={0.4in 0in 0in 0in},
        clip=false]{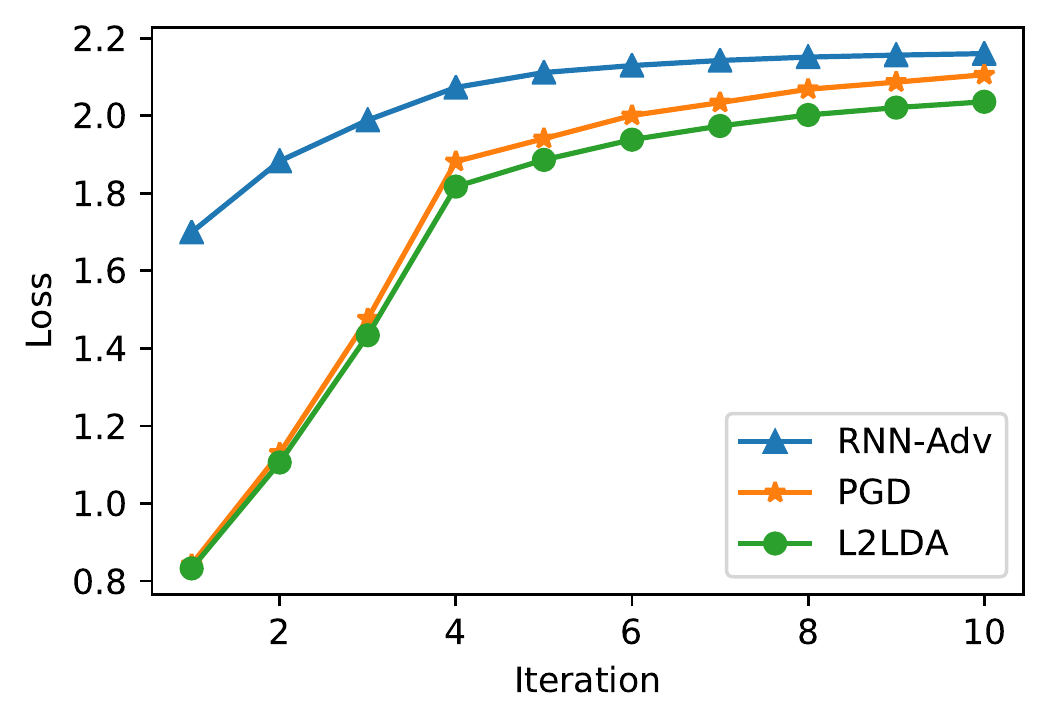}
        \caption{RNN-Adv} 
        \label{fig:bianrycontrolb}
    \end{subfigure}
    \begin{subfigure}[ht]{0.45\textwidth}
        \centering
        \includegraphics[width=\textwidth, trim={0in 0in 0.4in 0in}, clip=false]{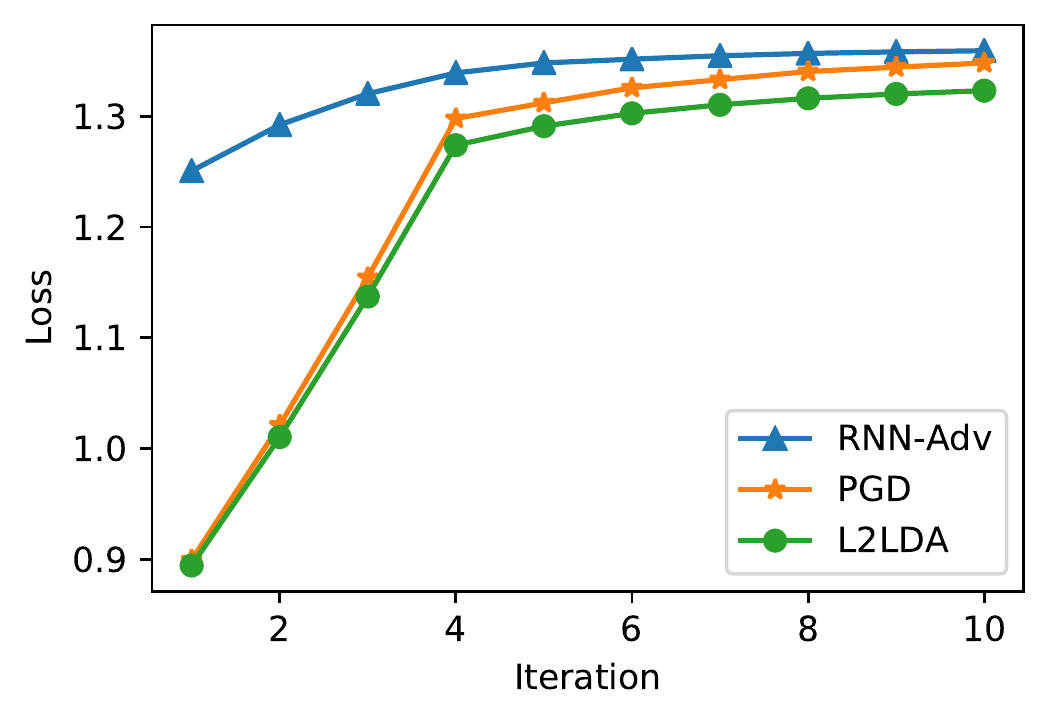}
        \caption{RNN-TRADES} 
        \label{fig:bianrycontrold}
    \end{subfigure}
    
    \label{fig:binary}
    \caption{Comparison of optimization trajectories among various attack algorithms. We evaluate four defense mechanisms, AdvTrain TRADES, RNN-Adv and RNN-TRADES, under three attackers including PGD, L2LDA and our proposed RNN-Adv. All attackers conduct 10-step perturbing process}
    \label{fig:trajectory}
    \vspace{-10pt}
\end{figure}

\textbf{Learned Optimizer}. As mentioned in Section~\ref{subsec:white}, the optimizer learned from PGD-based adversarial training can be regarded as an special attacker. Thus, we primarily investigate the update trajectories of different attackers to obtain an in-depth understanding of our RNN optimizer. For VGG-16 models trained from four defense methods, three attacker are used to generate perturbations in 10 steps respectively and losses are recorded as shown in Figure~\ref{fig:trajectory}. %(\textcolor{red}{unified scale?})

We can see clearly from these four figures that the losses obtained from RNN-Adv are always larger than others within 10 iterations, reflecting stronger attacks produced by our proposed optimizer. Moreover, it should be noted that the loss gap between  RNN-Adv and other attackers is much more prominent at some very beginning iterations. This in fact demonstrates an advantage of the learning-to-learn framework that the optimizer can converge faster than hand-designed algorithms. 
% \vspace{-20pt}

\noindent\textbf{Generalization to more attack steps.} Although our learned RNN optimizer is only trained under 10 steps, we show that it can generalize to more steps as an attacker. From Table~\ref{tab:generalization}, we can observe that the attacker is capable of producing much stronger adversarial examples by extending its attack steps to 40. Performance of our attacker is even comparable with that of PGD-100, which further demonstrates the superiority of our proposed method.
% \begin{table}[htb!]
% 	\centering
% 	\caption{Comparion among different L2L-based methods}
% 	\label{tab:generalization}
% 	\centering
% 	{\footnotesize
% 	\begin{center}
% 		\begin{tabular}{c|c|c}
% 			\hline
%         % \toprule
%         \diagbox{Defense}{Surrogate} & Plain-Net & PGD-Net \\
%         \hline
%         % \midrule
%         PGD-10 & 79.94 & 62.57 \\
%         TRADES & 77.01 & 65.41 \\
%         L2LDA & 76.37 & 60.32 \\
%         RNN-PGD & 80.58 & 63.17  \\
%         RNN-TRADES & 79.54  & 67.09\\
% 			\hline
%         % \bottomrule
% 		\end{tabular}	
% 	\end{center}}
% 	%\vspace{-0.2in}
% \end{table}

% \textbf{Generalization to more attack steps.} As we mentioned before, we show our 

\subsection{Performance on Black-box Transfer Attacks}
%{\color{red}(Maybe say ``PErformance on Black-box Transfer Attacks'')}

We further test the robustness of the proposed defense method under transer attack. As suggested by~\cite{athalye2018obfuscated}, this can be served as a sanity check to see whether our defense leads to obfuscated gradients
and gives a false sense of model robustness. 
%
%Under the black-box attack setting, we have no access to the gradient information of target models. The most intuitive way to conduct such attacks is by exploiting the transferability of adversarial examples, which means that malicious perturbations produced from one model can also be effective to fool other different models. Thus, 
Following procedures in~\cite{athalye2018obfuscated}, we first train a surrogate model with the same architecture of the target model using a different random seed, and then generate adversarial examples from the surrogate model to attack the target model.

\begin{table*}
\begin{floatrow}
\capbtabbox{

		\begin{tabular}{c|c|c}
% 			\hline
        \toprule
        \diagbox{Defense}{Step} & 10 & 40 \\
        % \hline
        \midrule
        Plain & 0.43 & \textbf{0.03} \\
        \midrule
        AdvTrain & 41.07  & \textbf{40.70}  \\
        TRADES & 48.38  & \textbf{48.27}  \\
        \midrule
        L2LDA & 34.54  & \textbf{34.19}  \\
        \midrule
        RNN-Adv & 43.21 & \textbf{42.89}  \\
        RNN-TRADES & 49.49 & \textbf{49.28}\\

% 			\hline
        \bottomrule
		\end{tabular}

}{
 \caption{Generalization to more steps of learned optimizer}
 \label{tab:generalization}
}
\capbtabbox{
		\begin{tabular}{c|c|c}
% 			\hline
        \toprule
        \diagbox{Defense}{Surrogate} & Plain-Net & PGD-Net \\
        % \hline
        \midrule
        AdvTrain & 79.94 & 62.57 \\
        TRADES & 77.01 & 65.41 \\
        \midrule
        L2LDA & 76.37 & 60.32 \\
        \midrule
        RNN-Adv & \textbf{80.58} & 63.17  \\
        RNN-TRADES & 79.54  & $\mathbf{67.09}$\\

% 			\hline
        \bottomrule
		\end{tabular}
}{
 \caption{Robust accuracy under black-box attack settings}
 \label{tab:black}
}
\end{floatrow}
\end{table*}

Specifically, we choose VGG-16 models obtained from various defense algorithms as our target models. In the meanwhile, we train two surrogate models: one is Plain-Net with natural training and the other is PGD-Net with 10-step PGD-based adversarial training. Results  are presented in Table~\ref{tab:black}. We can observe that our method outperforms all other baselines, with RNN-PGD and RNN-TRADES standing out in defending attacks from Plain-Net and PGD-Net respectively. It suggests great resistance of our L2L defense to transfer attacks.

\subsection{Loss Landscape Exploration}
\begin{figure}[t] 
    \centering
    \begin{subfigure}[ht]{0.325\textwidth}
        \centering
        \includegraphics[width=\textwidth,
        clip=false]{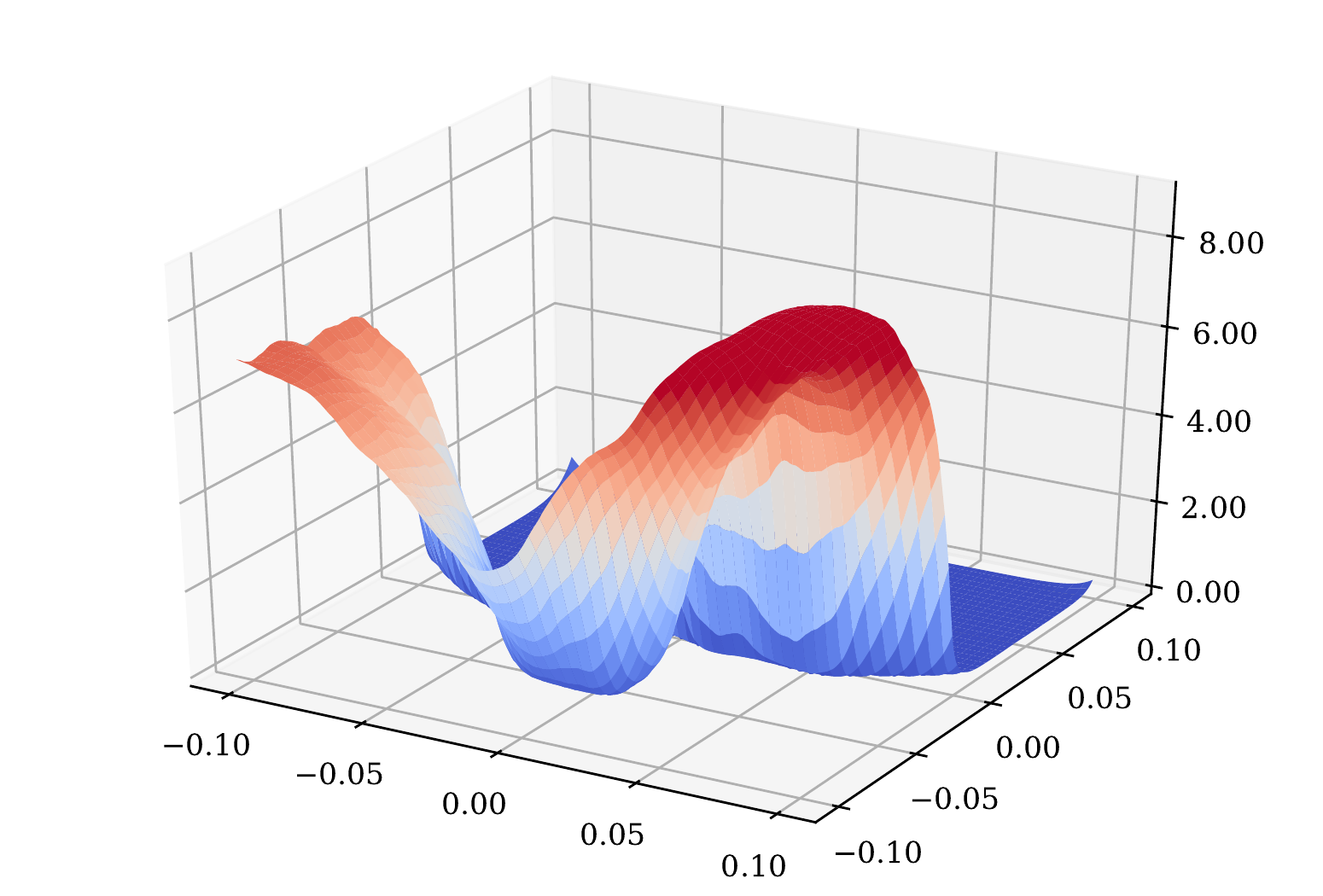}
        \caption{Plain} 
    \end{subfigure}
    \begin{subfigure}[ht]{0.325\textwidth}
        \centering
        \includegraphics[width=\textwidth, 
        clip=false]{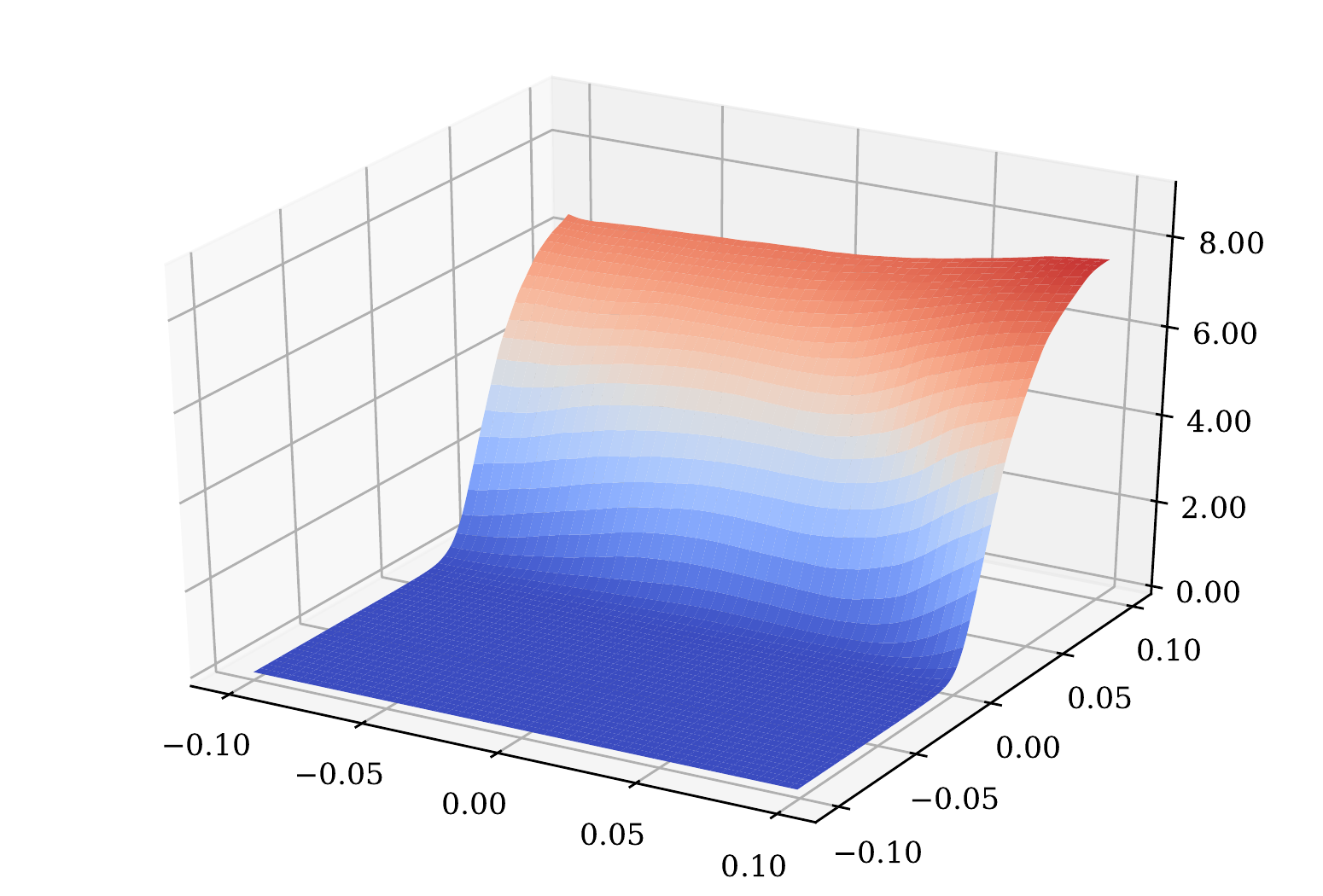}
        \caption{Adversarial Training}
        \label{fig:loss-adv}
    \end{subfigure}
    \begin{subfigure}[ht]{0.325\textwidth}
        \centering
        \includegraphics[width=\textwidth, 
        clip=false]{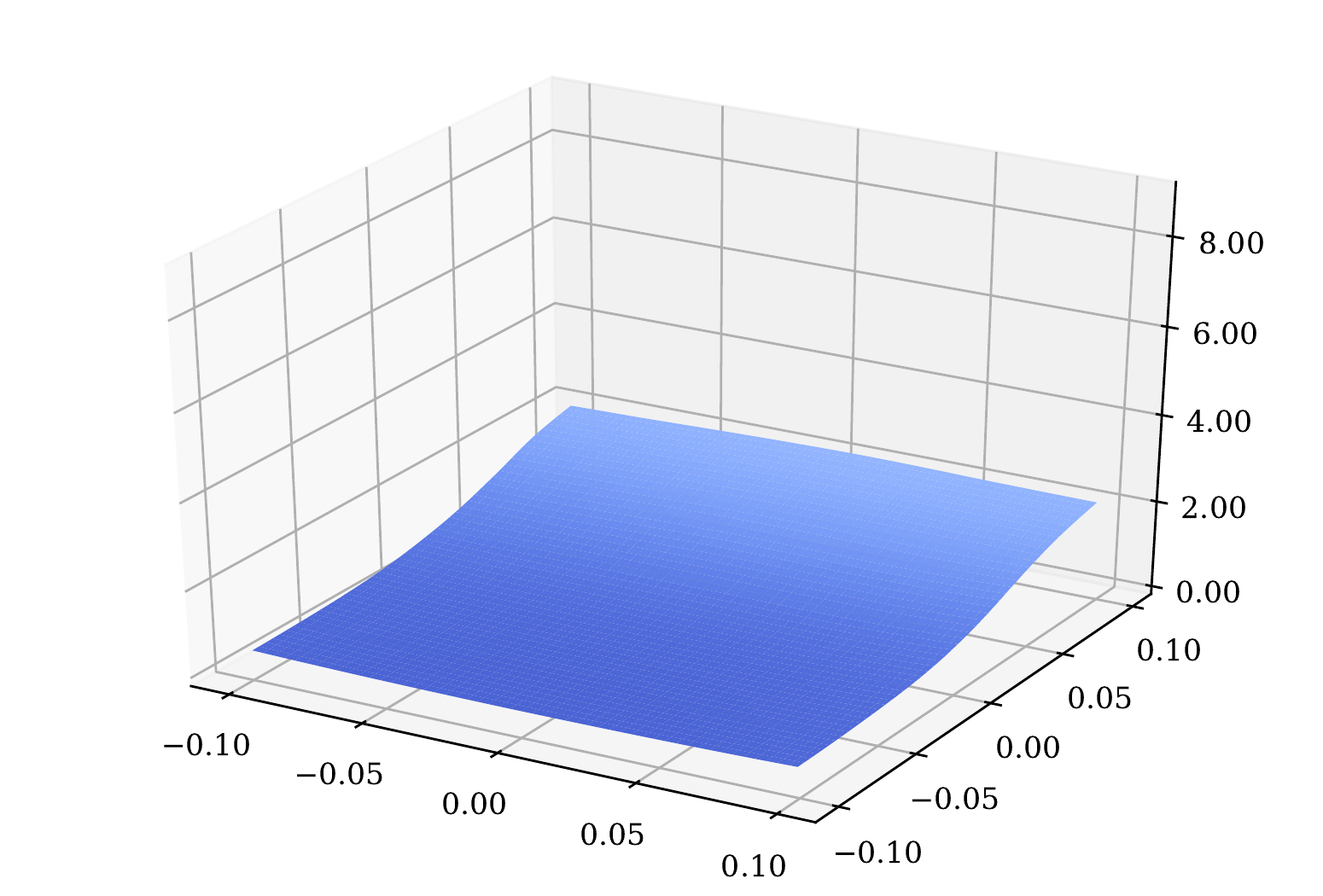}
        \caption{TRADES} 
        \label{fig:loss-trades}
    \end{subfigure}
    \begin{subfigure}[ht]{0.325\textwidth}
        \centering
        \includegraphics[width=\textwidth, clip=false]{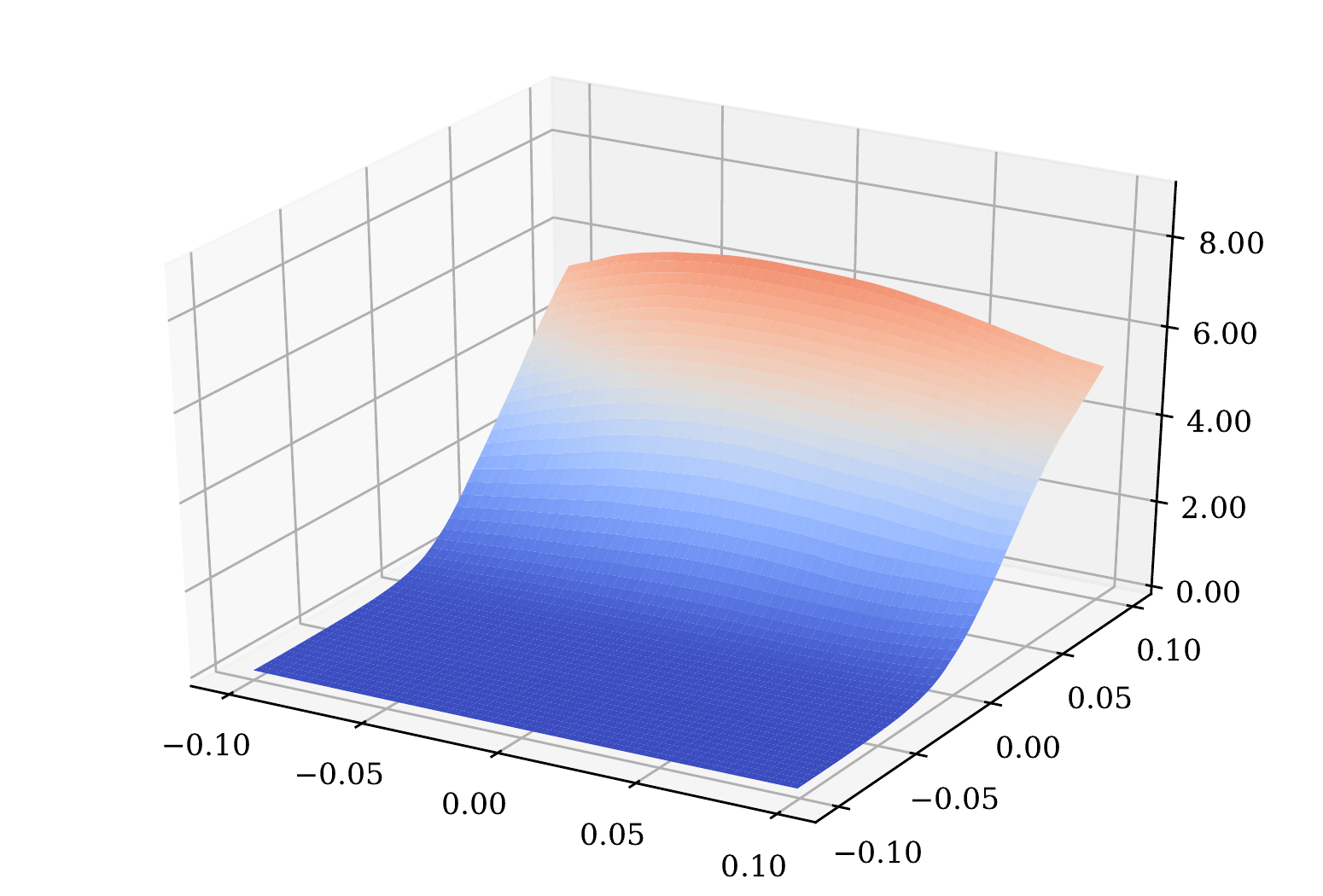}
        \caption{L2LDA} 
        \label{fig:loss-l2lda}
    \end{subfigure}
    \begin{subfigure}[ht]{0.325\textwidth}
        \centering
        \includegraphics[width=\textwidth, clip=false]{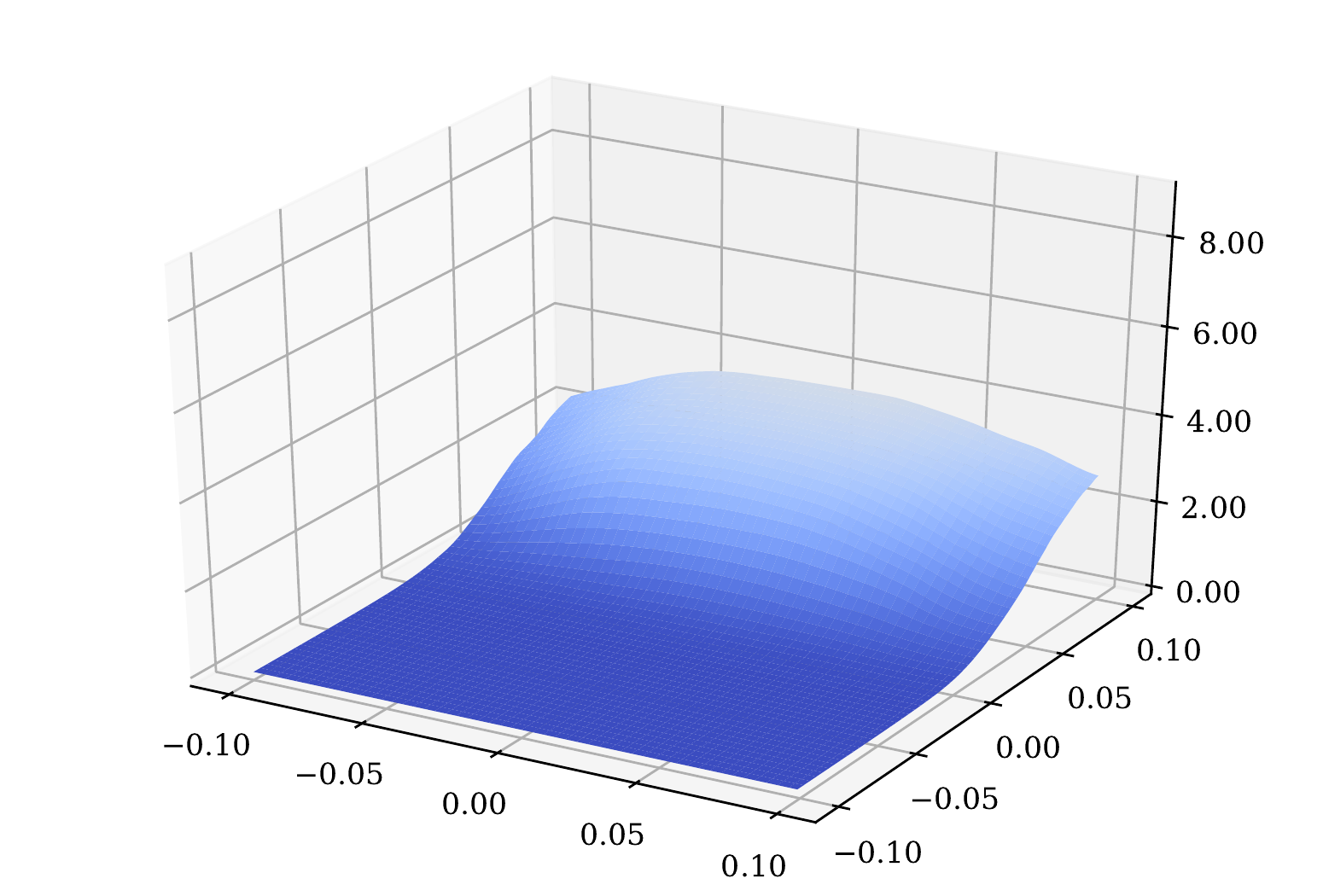}
        \caption{RNN-Adv} 
        \label{fig:loss-rnn-adv}
    \end{subfigure}
    \begin{subfigure}[ht]{0.325\textwidth}
        \centering
        \includegraphics[width=\textwidth, clip=false]{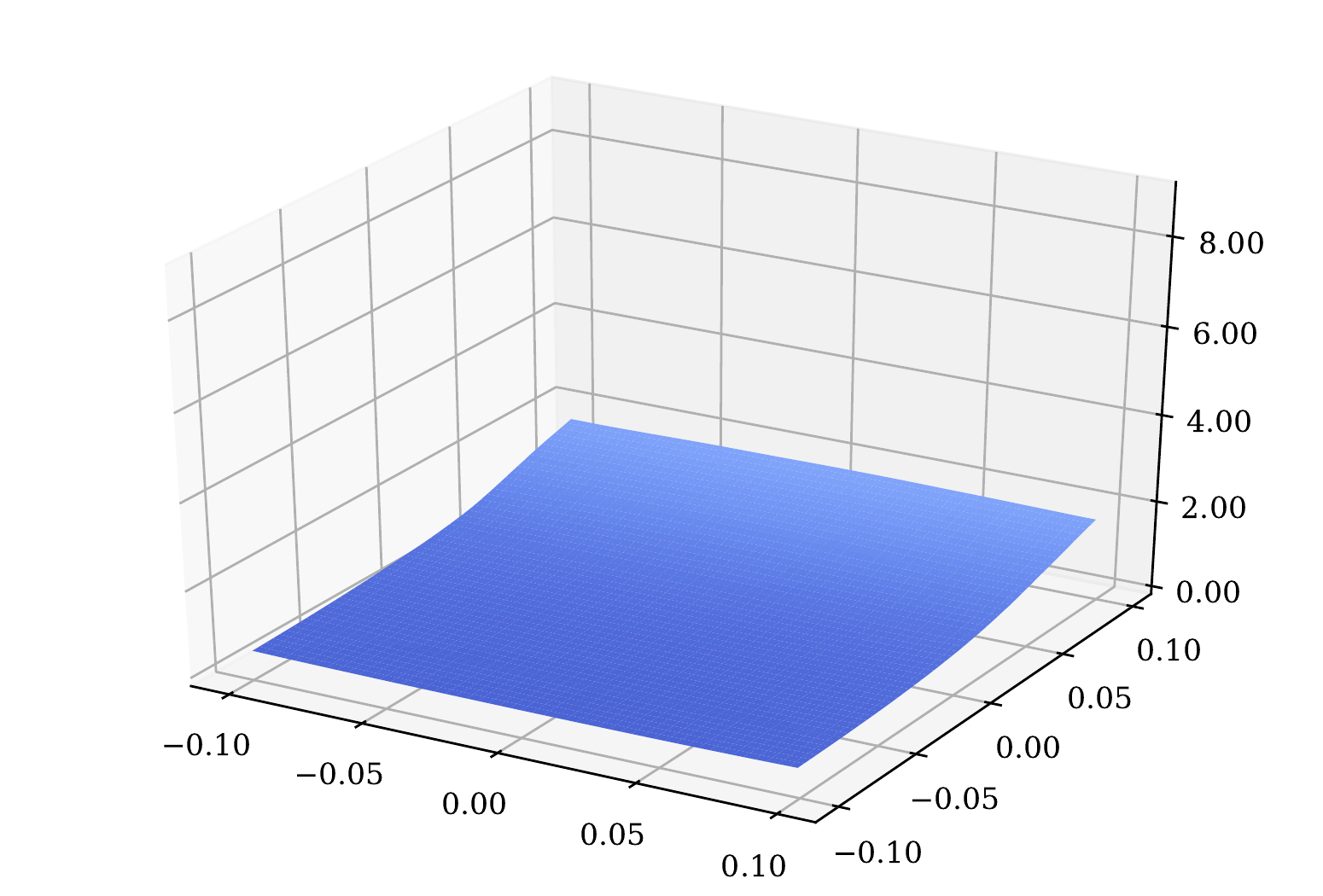}
        \caption{RNN-TRADES} 
        \label{fig:loss-rnn-trades}
    \end{subfigure}
    
    \caption{Comparison of loss landscapes among different training methods. %The $z$ axis represents the loss. If $\x$ is the original input, then we plot the loss varying along the space determined by two vectors: $r_1=\sign(\nabla_{\x} f(\x))$ and $r_2\sim Rademacher(0.5)$. Thus the following function is plotted: $z=\cL(x\cdot r_1 + y\cdot r_2)$. 
    The color gradually changes from blue~(low loss) to red~(high loss)}
    \label{fig:loss}
\end{figure}

To further verify the superior performance of the proposed algorithm,  we visualize the loss landscapes of VGG-16 models trained under different defense strategies, as shown in Figure~\ref{fig:loss}. According to the implementation in~\cite{engstrom2018evaluating}, we modify the input along a linear space defined by the sign of the gradient and a random Rademacher vector, where the x and y axes represent the magnitude of the perturbation added in each direction and the z axis represents the loss. It can be observed that loss surfaces of models trained from RNN-Adv and RNN-TRADES in Figure~\ref{fig:loss-rnn-adv} and \ref{fig:loss-rnn-trades} are much smoother than those of their counterparts in Figure~\ref{fig:loss-adv} and \ref{fig:loss-trades}. Besides, our method significantly reduces the loss value of perturbed data close to the original input. In particular, the maximum loss decreases roughly from $7.61$ in adversarial training to $3.66$ in RNN-Adv. Compared with L2LDA in Figure~\ref{fig:loss-l2lda}, the proposed RNN optimizer can contribute to less bumpier loss landscapes with smaller variance, which further demonstrates the stability and superiority of our L2L-based adversarial training.

% The results provide strong evidence for finding more robust models via SPROUT

%% file: ref.tex
\bibliographystyle{splncs04}
\bibliography{ref}

%% file: appendix.tex
\pagestyle{empty}
\appendix
% \section{Supplementary Material}
\section{Algorithm for TRADES}
\label{appendix:trades}
As we have emphasized, our proposed method can be incorporated into any adversarial training which can be formulated as a minimax optimization problem. Here we provide the detailed algorithm of RNN-TRADES in Algorithm~\ref{alg:rnn-trades}.

\begin{algorithm}[th]
 \caption{RNN-based TRADES}\label{alg:rnn-trades}
 \begin{algorithmic}[1]
 \State \textbf{Input}: clean data $\{(\x,\y)\}$, batch size $B$, step sizes $\alpha_1$ and $\alpha_2$, number of inner iterations $T$, classifier parameterized by $\theta$, RNN optimizer parameterized by $\phi$
 \State \textbf{Output}: Robust classifer $f_\theta$, learned optimizer $m_\phi$
 \State Randomly initialize $f_\theta$ and $m_\phi$, or initialize them with pre-trained configurations
 \Repeat
   \State Sample a mini-batch $M$ from clean data.
   \For{$(\x, \y)$ {\bfseries in} $B$}
   \State Initialization: $\bm{h}_0\leftarrow0$, $\cL_{\theta}\leftarrow 0$, $\cL_{\phi}\leftarrow 0$
   \State Gaussian augmentation: $\x_0' \leftarrow \x + 0.001\cdot\N(\bm{0},\bm{I})$
   \For{$t=0,\dots, T-1$} 
  \State $\bm{g}_t \leftarrow \nabla_{\x'}\cL(f_\theta(\x),f_\theta(\x_{t}'))$
  \State ${\bm{\delta}}_t,\bm{h}_{t+1} \leftarrow m_{\phi}(\bm{g}_t, \bm{h}_{t})$, where coordinate-wise update is applied
    \State $\x_{t+1}'\leftarrow \Pi_{\bbB(\x,\epsilon)}(\x_{t}'+\bm{\delta}_t)$
    \State $\cL_{\phi} \leftarrow \cL_{\phi} + w_{t+1}\cL(f_\theta(\x), f_\theta(\x'_{t+1})) $, where $w_{t+1}=t+1$
%   \ENDIF
   \EndFor
   \State $\cL_{\theta} \leftarrow \cL_{\theta}+\cL(f_\theta(\x),\y) + \cL(f_\theta(\x), f_\theta(\x'_{T}))/\lambda$ 
   \EndFor
   \State Update $\phi$ by $\phi \leftarrow \phi+\alpha_{1}\nabla_\phi\cL_{\phi}/B$
   \State Update $\theta$ by $\theta \leftarrow \theta - \alpha_{2}\nabla_\theta\cL_{\theta}/B$
 \Until{training converged}
 \end{algorithmic}

\end{algorithm}

\section{Additional Analysis}
\label{appendix:analysis}
\subsection{Time comparison}
In this section, we compared training time of our proposed methods with the original adversarial training. We still conduct analysis of VGG-16 on CIFAR-10 dataset. From results in Table~\ref{tab:moretime}, it can be observed that our methods approximately double the overall training time per epoch, which is not a heavy burden with improved performance taken into account. 
%To eliminate the influence of training time, we then increase iterations of adversarial training from 10 to 20 for fair comparison. 
\begin{table}[]
    \centering
		\begin{tabular}{c|cc}
% 			\hline
        \toprule
        & Training Time & Ratio of RNN counterpart\\
        \midrule
        AdvTrain & 122.24 & 2.20 (268.50)  \\
        TRADES & 189.43 & 2.34 (443.52) \\
        % \midrule
        % RNN-Adv & 268.50 \\
        % RNN-TRADES & 443.52 \\
        \bottomrule
		\end{tabular}
    \caption{Time comparison with original adversarial training. Here we report the ratio of our proposed method to its original counterpart, for example, $T_{\text{RNN-Adv}}/T_{\text{AdvTrain}}$. In the parentheses, we report the training time per epoch of our proposed method including RNN-Adv and RNN-TRADES}
    \label{tab:moretime}
\end{table}

% \begin{table*}
% \begin{floatrow}
% \capbtabbox{

% 		\begin{tabular}{c|c}
% % 			\hline
%         \toprule
%         & Training Time\\
%         \midrule
%         AdvTrain & 122.24 \\
%         TRADES & 189.43 \\
%         \midrule
%         RNN-Adv & \\
%         RNN-TRADES & \\

% % 			\hline
%         \bottomrule
% 		\end{tabular}

% }{
%  \caption{Time}
%  \label{tab:xxx}
% }
% \capbtabbox{
% 		\begin{tabular}{c|c|c}
% % 			\hline
%         \toprule
%         \diagbox{Defense}{Attack} & Plain & PGD-10 \\
%         % \hline
%         \midrule
%         AdvTrain &  &  \\
%         AdvTrain-20 & 80.06 & 43.98 \\
%         RNN-Adv &  &  \\

% % 			\hline
%         \bottomrule
% 		\end{tabular}
% }{
%  \caption{Robust accuracy under black-box attack settings}
%  \label{tab:x}
% }
% \end{floatrow}
% \end{table*}
\subsection{Trajectory}
\begin{figure}[t]
    \centering
    \includegraphics[width=0.5\textwidth]{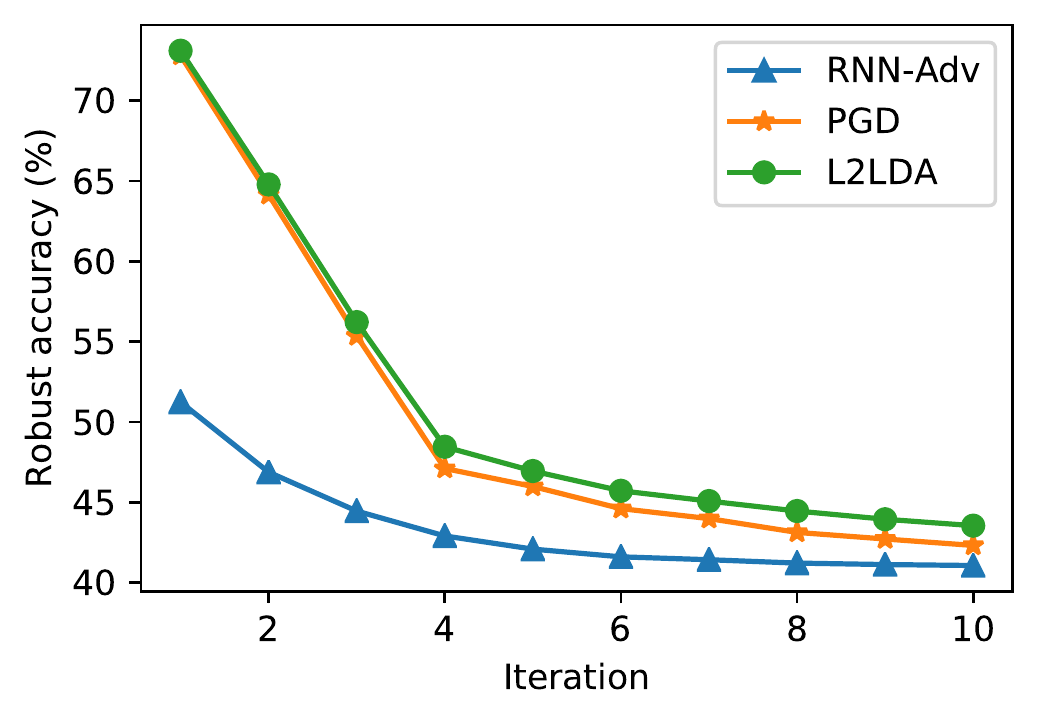}
    \caption{Accuracy trajectory}
    \label{fig:acc}
\end{figure}

A similar trajectory can be observed in terms of classification accuracy as well when the model is attacked by different attackers. In Figure~\ref{fig:acc}, the robust accuracy under RNN-Adv drops most rapidly and also achieves the lowest point after the entire 10-step attack. It further verifies that our learned optimizer can guide the optimization along a better trajectory for the inner problem, meaning that crafted adversarial examples are much more powerful. This in turn contributes to a more robust model.

\begin{table}[h]
	\centering
	\caption{Robust accuracy under white-box attacks (Rest. ImageNet, ResNet18)}
    \label{tab:imagenet}
	\centering
	{\footnotesize
	\begin{center}
		\begin{tabular}{c|cccc||c}
% 			\hline
\toprule
            \diagbox{Defense}{Attack} & Natural &  PGD-10 & PGD-100 & CW100 & Min \\
            % \hline
            \midrule
            Plain & \textbf{97.66} & 0.00 & 0.00  & 0.00 & 0.00 \\
            \midrule
            AdvTrain & 91.43 &  8.50 & 4.88   & 9.04   & 4.88  \\
            % CW100 & & & & & &  \\
            %TRADES & 99.52  & & 95.77 & & & 95.64 & \\
            TRADES & 72.51   & 6.96 & 4.80  & 10.47   &  4.80   \\
            \midrule
            RNN-Adv & 90.28  & \textbf{10.13} & \textbf{6.04} & 12.98   & \textbf{6.04}  \\
            RNN-TRADES & 73.84   & 9.76  & 5.79 & \textbf{13.22}  & 5.79 \\
           % LSTM-Trades & & & & & & & \\
% 			\hline
\bottomrule
		\end{tabular}	
	\end{center}}
	%\vspace{-0.2in}
\end{table}

\section{Experiments on Restricted ImageNet}
\label{appendix:imagenet}
Without loss of generality, we conduct extra experimental analysis on a larger scale dataset, Restricted ImageNet~\cite{tsipras2018robustness}. It is a subset of 9 different super-classes extracted from the entire ImageNet to reduce the computational burden for adversarial training. We adopt the structure of ResNet-18 as the classifier for this dataset. Following the literature~\cite{sinha2019harnessing,tsipras2018robustness}, the inner attack strength of all defense methods is set to be 0.005 in $l_\infty$ ball while models trained from these mechanisms are evaluated under the attack with the radius of 0.025. Note that in this experiment we only fine-tune the model starting from the naturally trained one for 2 epochs to observe different performance of defense strategies. 

From Table~\ref{tab:imagenet}, we can clearly see that our proposed methods such as RNN-Adv and RNN-TRADES consistently improve the robust accuracy compared with their original adversarial training algorithms. Specifically, models trained by our methods are always $2\%-3\%$ percents more robust that others under various attacks.

%% file: main.bbl
\begin{thebibliography}{10}
\providecommand{\url}[1]{\texttt{#1}}
\providecommand{\urlprefix}{URL }
\providecommand{\doi}[1]{https://doi.org/#1}

\bibitem{andrychowicz2016learning}
Andrychowicz, M., Denil, M., Gomez, S., Hoffman, M.W., Pfau, D., Schaul, T.,
  Shillingford, B., De~Freitas, N.: Learning to learn by gradient descent by
  gradient descent. In: Advances in neural information processing systems. pp.
  3981--3989 (2016)

\bibitem{athalye2018obfuscated}
Athalye, A., Carlini, N., Wagner, D.: Obfuscated gradients give a false sense
  of security: Circumventing defenses to adversarial examples. In:
  International Conference on Machine Learning. pp. 274--283 (2018)

\bibitem{brendel2017decision}
Brendel, W., Rauber, J., Bethge, M.: Decision-based adversarial attacks:
  Reliable attacks against black-box machine learning models. arXiv preprint
  arXiv:1712.04248  (2017)

\bibitem{carlini2017towards}
Carlini, N., Wagner, D.: Towards evaluating the robustness of neural networks.
  In: 2017 ieee symposium on security and privacy (sp). pp. 39--57. IEEE (2017)

\bibitem{cheng2018query}
Cheng, M., Le, T., Chen, P.Y., Yi, J., Zhang, H., Hsieh, C.J.: Query-efficient
  hard-label black-box attack: An optimization-based approach. arXiv preprint
  arXiv:1807.04457  (2018)

\bibitem{cheng2020signopt}
Cheng, M., Singh, S., Chen, P.H., Chen, P.Y., Liu, S., Hsieh, C.J.: Sign-opt: A
  query-efficient hard-label adversarial attack. In: ICLR (2020)

\bibitem{cotter1990fixed}
Cotter, N.E., Conwell, P.R.: Fixed-weight networks can learn. In: 1990 IJCNN
  International Joint Conference on Neural Networks. pp. 553--559. IEEE (1990)

\bibitem{elsken2018neural}
Elsken, T., Metzen, J.H., Hutter, F.: Neural architecture search: A survey.
  arXiv preprint arXiv:1808.05377  (2018)

\bibitem{engstrom2018evaluating}
Engstrom, L., Ilyas, A., Athalye, A.: Evaluating and understanding the
  robustness of adversarial logit pairing. arXiv preprint arXiv:1807.10272
  (2018)

\bibitem{finn2017model}
Finn, C., Abbeel, P., Levine, S.: Model-agnostic meta-learning for fast
  adaptation of deep networks. In: Proceedings of the 34th International
  Conference on Machine Learning-Volume 70. pp. 1126--1135. JMLR. org (2017)

\bibitem{goodfellow2014explaining}
Goodfellow, I.J., Shlens, J., Szegedy, C.: Explaining and harnessing
  adversarial examples. arXiv preprint arXiv:1412.6572  (2014)

\bibitem{hendrycks2019natural}
Hendrycks, D., Zhao, K., Basart, S., Steinhardt, J., Song, D.: Natural
  adversarial examples. arXiv preprint arXiv:1907.07174  (2019)

\bibitem{jang2019adversarial}
Jang, Y., Zhao, T., Hong, S., Lee, H.: Adversarial defense via learning to
  generate diverse attacks. In: Proceedings of the IEEE International
  Conference on Computer Vision. pp. 2740--2749 (2019)

\bibitem{jiang2018learning}
Jiang, H., Chen, Z., Shi, Y., Dai, B., Zhao, T.: Learning to defense by
  learning to attack. arXiv preprint arXiv:1811.01213  (2018)

\bibitem{krizhevsky2010cifar}
Krizhevsky, A., Nair, V., Hinton, G.: Cifar-10 (canadian institute for advanced
  research). URL http://www. cs. toronto. edu/kriz/cifar. html  \textbf{8}
  (2010)

\bibitem{kurakin2016adversarial2}
Kurakin, A., Goodfellow, I., Bengio, S.: Adversarial examples in the physical
  world. arXiv preprint arXiv:1607.02533  (2016)

\bibitem{kurakin2016adversarial}
Kurakin, A., Goodfellow, I., Bengio, S.: Adversarial machine learning at scale.
  arXiv preprint arXiv:1611.01236  (2016)

\bibitem{lecun1998gradient}
LeCun, Y., Bottou, L., Bengio, Y., Haffner, P.: Gradient-based learning applied
  to document recognition. Proceedings of the IEEE  \textbf{86}(11),
  2278--2324 (1998)

\bibitem{lee2017generative}
Lee, H., Han, S., Lee, J.: Generative adversarial trainer: Defense to
  adversarial perturbations with gan. arXiv preprint arXiv:1705.03387  (2017)

\bibitem{lv2017learning}
Lv, K., Jiang, S., Li, J.: Learning gradient descent: Better generalization and
  longer horizons. In: Proceedings of the 34th International Conference on
  Machine Learning-Volume 70. pp. 2247--2255. JMLR. org (2017)

\bibitem{madry2017towards}
Madry, A., Makelov, A., Schmidt, L., Tsipras, D., Vladu, A.: Towards deep
  learning models resistant to adversarial attacks. arXiv preprint
  arXiv:1706.06083  (2017)

\bibitem{metz2019understanding}
Metz, L., Maheswaranathan, N., Nixon, J., Freeman, D., Sohl-Dickstein, J.:
  Understanding and correcting pathologies in the training of learned
  optimizers. In: International Conference on Machine Learning. pp. 4556--4565
  (2019)

\bibitem{moosavi2016deepfool}
Moosavi-Dezfooli, S.M., Fawzi, A., Frossard, P.: Deepfool: a simple and
  accurate method to fool deep neural networks. In: Proceedings of the IEEE
  conference on computer vision and pattern recognition. pp. 2574--2582 (2016)

\bibitem{nocedal2006numerical}
Nocedal, J., Wright, S.: Numerical optimization. Springer Science \& Business
  Media (2006)

\bibitem{reddy2018nag}
Reddy~Mopuri, K., Ojha, U., Garg, U., Venkatesh~Babu, R.: Nag: Network for
  adversary generation. In: Proceedings of the IEEE Conference on Computer
  Vision and Pattern Recognition. pp. 742--751 (2018)

\bibitem{ruan2019learning}
Ruan, Y., Xiong, Y., Reddi, S., Kumar, S., Hsieh, C.J.: Learning to learn by
  zeroth-order oracle. arXiv preprint arXiv:1910.09464  (2019)

\bibitem{samangouei2018defense}
Samangouei, P., Kabkab, M., Chellappa, R.: Defense-gan: Protecting classifiers
  against adversarial attacks using generative models. arXiv preprint
  arXiv:1805.06605  (2018)

\bibitem{simonyan2014very}
Simonyan, K., Zisserman, A.: Very deep convolutional networks for large-scale
  image recognition. arXiv preprint arXiv:1409.1556  (2014)

\bibitem{sinha2019harnessing}
Sinha, A., Singh, M., Kumari, N., Krishnamurthy, B., Machiraju, H.,
  Balasubramanian, V.N.: Harnessing the vulnerability of latent layers in
  adversarially trained models. arXiv preprint arXiv:1905.05186  (2019)

\bibitem{tsipras2018robustness}
Tsipras, D., Santurkar, S., Engstrom, L., Turner, A., Madry, A.: Robustness may
  be at odds with accuracy. arXiv preprint arXiv:1805.12152  (2018)

\bibitem{wang2019direct}
Wang, H., Yu, C.N.: A direct approach to robust deep learning using adversarial
  networks. arXiv preprint arXiv:1905.09591  (2019)

\bibitem{wang2019bilateral}
Wang, J., Zhang, H.: Bilateral adversarial training: Towards fast training of
  more robust models against adversarial attacks. In: Proceedings of the IEEE
  International Conference on Computer Vision. pp. 6629--6638 (2019)

\bibitem{wichrowska2017learned}
Wichrowska, O., Maheswaranathan, N., Hoffman, M.W., Colmenarejo, S.G., Denil,
  M., de~Freitas, N., Sohl-Dickstein, J.: Learned optimizers that scale and
  generalize. In: Proceedings of the 34th International Conference on Machine
  Learning-Volume 70. pp. 3751--3760. JMLR. org (2017)

\bibitem{wu2018understanding}
Wu, Y., Ren, M., Liao, R., Grosse, R.: Understanding short-horizon bias in
  stochastic meta-optimization. arXiv preprint arXiv:1803.02021  (2018)

\bibitem{xiao2018generating}
Xiao, C., Li, B., Zhu, J.Y., He, W., Liu, M., Song, D.: Generating adversarial
  examples with adversarial networks. In: Proceedings of the 27th International
  Joint Conference on Artificial Intelligence. pp. 3905--3911 (2018)

\bibitem{xie2019feature}
Xie, C., Wu, Y., Maaten, L.v.d., Yuille, A.L., He, K.: Feature denoising for
  improving adversarial robustness. In: Proceedings of the IEEE Conference on
  Computer Vision and Pattern Recognition. pp. 501--509 (2019)

\bibitem{younger2001meta}
Younger, A.S., Hochreiter, S., Conwell, P.R.: Meta-learning with
  backpropagation. In: IJCNN'01. International Joint Conference on Neural
  Networks. Proceedings (Cat. No. 01CH37222). vol.~3. IEEE (2001)

\bibitem{zagoruyko2016wide}
Zagoruyko, S., Komodakis, N.: Wide residual networks. arXiv preprint
  arXiv:1605.07146  (2016)

\bibitem{zhang2019theoretically}
Zhang, H., Yu, Y., Jiao, J., Xing, E.P., Ghaoui, L.E., Jordan, M.I.:
  Theoretically principled trade-off between robustness and accuracy. arXiv
  preprint arXiv:1901.08573  (2019)

\end{thebibliography}
